\documentclass{article}

\usepackage{microtype}
\usepackage{graphicx}
\usepackage{subcaption}
\usepackage{booktabs}
\usepackage{tcolorbox}
\usepackage{threeparttable}
\usepackage{pifont}

\usepackage{booktabs}
\usepackage{tabularx}
\usepackage{array}
\usepackage{enumitem}
\usepackage{xurl}  
\usepackage{hyperref}

\usepackage[preprint]{icml2026}

\usepackage{amsmath}
\usepackage{amssymb}
\usepackage{mathtools}
\usepackage{amsthm}
\usepackage{csquotes}

\theoremstyle{plain}

\theoremstyle{definition}

\theoremstyle{remark}

\usepackage[textsize=tiny]{todonotes}

\icmltitlerunning{Achieving Time Series Reasoning Requires Rethinking Model Design, Tasks Formulation, and Evaluation}

\begin{document}

\twocolumn[
  \icmltitle{Achieving Time Series Reasoning Requires Rethinking Model Design, Tasks Formulation, and Evaluation}

\icmlsetsymbol{equal}{$\star$}
\icmlsetsymbol{corresponding}{$\dagger$}

\begin{icmlauthorlist}
\icmlauthor{Yaxuan Kong}{Oxford,equal}
\icmlauthor{Yiyuan Yang}{Oxford,equal}
\icmlauthor{Shiyu Wang}{Ant}
\icmlauthor{Chenghao Liu}{Salesforce}
\icmlauthor{Yuxuan Liang}{HKUST}
\icmlauthor{Ming Jin}{Griffith,corresponding} \\
\icmlauthor{Stefan Zohren}{Oxford}
\icmlauthor{Dan Pei}{THU}
\icmlauthor{Yan Liu}{USC}
\icmlauthor{Qingsong Wen}{Oxford,Squirrel,corresponding}
\end{icmlauthorlist}

\icmlaffiliation{Oxford}{University of Oxford, UK}
\icmlaffiliation{Ant}{Ant Group, China}
\icmlaffiliation{Salesforce}{Salesforce, Singapore}
\icmlaffiliation{HKUST}{Hong Kong University of Science and Technology (Guangzhou), China}
\icmlaffiliation{Griffith}{Griffith University, Australia}
\icmlaffiliation{THU}{Tsinghua University, China}
\icmlaffiliation{USC}{University of Southern California, USA}
\icmlaffiliation{Squirrel}{Squirrel Ai Learning, USA}

\icmlcorrespondingauthor{Ming Jin}{mingjinedu@gmail.com}
\icmlcorrespondingauthor{Qingsong Wen}{qingsongedu@gmail.com}

\icmlkeywords{Time Series Reasoning, Multimodal LLMs}

\vskip 0.3in
]

\printAffiliationsAndNotice{\icmlEqualContribution}

\begin{abstract}
Understanding time series data is fundamental to many real-world applications. Recent work explores multimodal large language models (MLLMs) to enhance time series understanding with contextual information beyond numerical signals. This area has grown from 7 papers in 2023 to over 580 in 2025, yet existing methods struggle in real-world settings. We analyze 20 influential works from 2025 across model design, task formulation, and evaluation, and identify critical gaps: methods adapt NLP techniques with limited attention to core time series properties; tasks remain restricted to traditional prediction and classification; and evaluations emphasize benchmarks over robustness, interpretability, or decision relevance. \textbf{We argue that achieving time series reasoning requires rethinking model design, task formulation, and evaluation together.} We define time series reasoning, outline challenges and future directions, and call on researchers to develop unified frameworks for robust, interpretable, and decision-relevant reasoning in real-world applications. The material is available at \url{https://github.com/Eleanorkong/Awesome-Time-Series-Reasoning}.
\end{abstract}

\section{Introduction}
Time series analysis has long been a cornerstone of real-world applications across domains such as finance, healthcare, and energy~\cite{nie2022time, yang2023dcdetector, kong2025fusing}. With the rise of large language models (LLMs) and multimodal large language models (MLLMs), researchers have begun to explore how contextual information beyond numerical data can enhance time series understanding and decision-making~\cite{jin2024time, liu2025picture, wang2025itformer, pmlr-v267-li25ah}. This shift has gained remarkable momentum, with papers on time series reasoning growing from 7 in 2023 to 69 in 2024 and exceeding 580 in 2025\footnote{We collected papers via systematic search and provided an interactive dashboard for publication trends in Appendix~\ref{appendix_related_works}.}. Researchers increasingly adopt techniques from the natural language processing (NLP) domain, such as chain-of-thought prompts, instruction tuning, and agentic workflows, to advance time series analysis~\cite{jin2024position, zhang2025timemaster, li2025zara, wang2026slep, lan2025gem}. We believe that time series reasoning represents an essential step toward time series intelligence, where insights from analysis directly inform decision-making. This leads us to ask a fundamental question: \textit{Has time series reasoning truly achieved this potential, and can we now reliably use it for real-world decision-making?}

\begin{figure}[!t]
\begin{center}
\includegraphics[width = 1\linewidth]{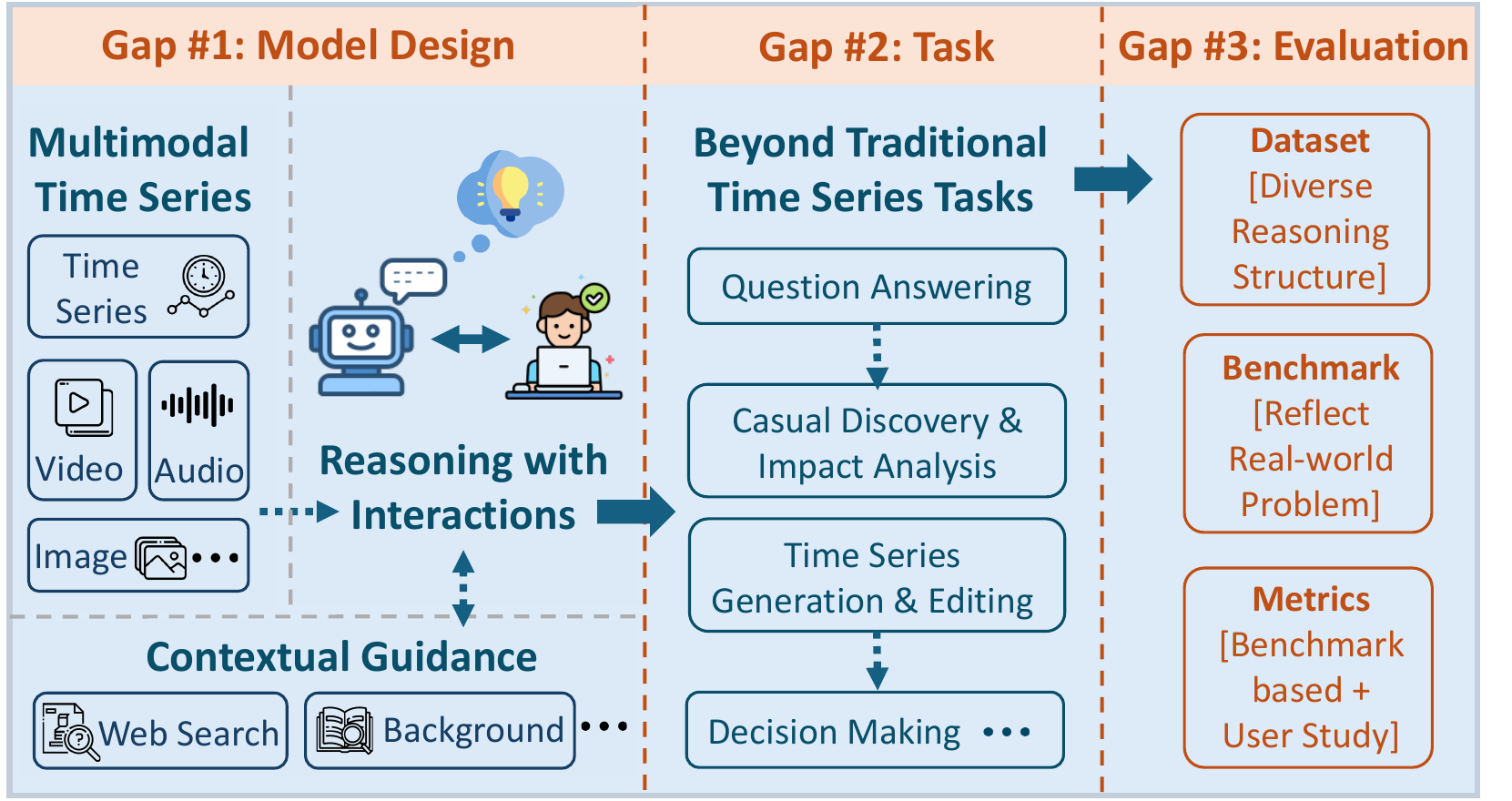}
\end{center}
\vspace{-1mm}
\caption{Three Gaps in Achieving Time Series Reasoning.}
\label{Figure_1}
\vspace{-5mm}
\end{figure}

A concrete case study highlights why this question matters. Consider a financial analyst asked to explain why a stock dropped $\sim$15\% over the past week and to recommend an action. The analyst must \textbf{(i)} understand temporal patterns in price data including trends, volatility, and volume, \textbf{(ii)} retrieve and synthesize relevant market news and company filings such as earnings guidance, macro announcements, and regulatory headlines, \textbf{(iii)} reason through causal relationships between news events and price movements to distinguish sector-wide drawdowns from firm-specific events, and \textbf{(iv)} verify the explanation against evidence with correct time alignment. This task seems straightforward, yet current methods struggle with open-ended decision-making questions, miss critical external drivers, show limited understanding of time series characteristics, and lack verification mechanisms. As a result, they cannot synthesize these factors into a coherent causal narrative that would allow an analyst to confidently recommend whether to buy, hold, or exit the position\footnote{Details about background, task and insights are available in Appendix~\ref{appendix_case_study}.}. This example forces a sharper question: \textit{What is still missing in current time series reasoning research that blocks real-world, evidence-grounded decision-making?}

To understand this gap, we analyzed 20 influential papers from 2025\footnote{Details are available in Appendix~\ref{appendix_related_works}.} and assessed each along three dimensions: (1) \textbf{Model Design}. How do models represent time series characteristics through raw sequences or encoded embeddings? How is external knowledge integrated through static prompts, dynamic retrieval, or cross-modal fusion? How is reasoning performed through chain-of-thought (CoT), instruction tuning, or agentic workflows? How do models refine and validate outputs? (2) \textbf{Task Formulation}. What specific tasks are these methods designed to solve, and do they extend beyond traditional forecasting and classification? (3) \textbf{Evaluation}. How is performance measured? Do evaluations assess reasoning quality, explanation coherence, and real-world applicability beyond benchmark accuracy?

Our analysis reveals several gaps that explain the case study failures. \textbf{Gap 1: Model Design}.  Most research adapts NLP techniques such as prompting, instruction tuning, and preference optimization. Contextual guidance is limited to static inputs without dynamic retrieval, time alignment, or verification loops. Many methods also serialize time series directly, raising questions about whether LLMs truly capture temporal characteristics beyond surface correlations. \textbf{Gap 2: Task Formulation}. Current tasks such as next-value prediction, pattern classification, and templated Question Answering (QA) are well-defined but narrow. They do not usually address causal analysis, multimodal integration, and decision-making under uncertainty that real-world problems demand. \textbf{Gap 3: Evaluation}. Methods are assessed on benchmarks or auto-generated QA that emphasize accuracy and text similarity. They rarely test scenarios that require multi-step reasoning, evidence verification, or handling of novel situations such as conflicting information. We observed that benchmark progress does not reliably translate into decision-ready reasoning for real-world applications.

Based on these observations, \textbf{we argue that achieving time series reasoning requires fundamentally rethinking model design, task formulation, and evaluation together}.  To support this claim, we first define time series reasoning, summarize its major reasoning types, and clarify how it relates to time series tasks. We then revisit the gaps identified above. (1) \textbf{Model Design.} We discuss the components needed for reliable reasoning, including understanding time series characteristics, contextual guidance, reasoning process, and iterative feedback. (2) \textbf{Task Formulation.} We show how time series reasoning enables broader problem settings, including question answering, causal inference and impact analysis, and time series generation and editing. (3) \textbf{Evaluation.} We review existing resources and advocate evaluations beyond benchmark accuracy that assess multi-step reasoning, evidence grounding and verification, robustness to conflicting information, and decision relevance. Finally, we call on researchers to develop unified frameworks that align these three dimensions toward robust, interpretable, and decision-ready time series reasoning.

\begin{table*}[!t]
    \caption{Comparison of reasoning types by task objective, including definitions and example applications in time series analysis.}
    \centering
    \small
    \resizebox{\linewidth}{!}{
        \renewcommand{\arraystretch}{1}
        \begin{tabular}{p{3.5cm}p{6.5cm}p{7.5cm}}
            \toprule
            \textbf{Reasoning Type} & \textbf{Definition} & \textbf{Example} \\
            \midrule
            Deductive Reasoning & Derives specific, logically consistent conclusions from general principles or rules. & Confirming a 25\% sales spike aligns with a known rule that holiday promotions increase sales by 20--30\%. \\
            \midrule
            Inductive Reasoning & Infers general rules or patterns from specific observations. & Hypothesizing that sales spikes every December indicate seasonality. \\
            \midrule
            Etiological Reasoning & Identifies and explains the underlying causes or mechanisms behind observed patterns. & Explaining a sudden drop in energy consumption by analyzing potential causes. \\
            \midrule
            Causal Reasoning & Measures causal effect relationships and requires a clear demonstration of true causal impacts beyond etiological reasoning. & Evaluating if a new pricing strategy caused sustained sales increases by comparing pre- and post-policy data. \\
            \midrule
            Analogical Reasoning & Identifies similarities between different time series or contexts. & Comparing sales patterns across stores to infer shared seasonality. \\
            \midrule
            Counterfactual Reasoning & Explores ``what-if'' scenarios by imagining alternative conditions or interventions. & Analyzing what sales would look like without a promotional discount in June. \\
            \midrule
            Mathematical Reasoning & Applies mathematical techniques or logical derivations to interpret data. & Proving stationarity by examining bounded partial sums over time. \\
            \midrule
            Abductive Reasoning & Forms the most plausible explanation for observations when evidence is incomplete. & Proposing a supply chain issue as the cause for an unexplained sales dip. \\
            \bottomrule
        \end{tabular}
    }
    \label{tab:additional_reasoning_types}
    \vspace{-5pt}
\end{table*}

\section{Time Series Reasoning}
\label{section_2}
\subsection{What is Time Series Data?}
Time series data consists of observations indexed by time, describing how a variable or system evolves temporally. A \textit{univariate} time series is a sequence of scalar measurements $X=\{x_1,x_2,\ldots,x_T\}\in\mathbb{R}^{T}$, while a \textit{multivariate} time series records an $N$-dimensional vector at each step, written as $X\in\mathbb{R}^{N\times T}$. Time series may be regularly or irregularly sampled, where irregularity often introduces missing values and requires modeling time gaps. Moreover, in many real-world multivariate settings, variables exhibit structured dependencies such as spatial or relational links, often called \textit{spatio-temporal} data. This motivates graph-based representations viewing data as snapshots $\mathcal{G}=\{G_1,\ldots,G_T\}$, where each $G_t=(A_t,X_t)$ comprises an adjacency matrix $A_t\in\mathbb{R}^{N\times N}$ encoding variable relations and a feature matrix $X_t\in\mathbb{R}^{N\times D}$ encoding node features~\cite{jin2024position}.

\subsection{What is Time Series Reasoning?}
Time series reasoning denotes the ability of an MLLM to process and interpret time series data with human-like logic~\cite{jin2024time, chang2025survey, ning2025towards}. Rather than mapping a sequence directly to a single output, it identifies temporal patterns such as trends, seasonality, and regime shifts, and integrates this evidence with contextual information. The model then reasons through this evidence and refines its conclusions to produce grounded natural language explanations. Unlike task-specific models, this approach provides a unified interface that generalizes across domains while it keeps reasoning interpretable. We view time series reasoning as essential to time series intelligence, where models understand why patterns occur, explain what they mean, and inform real-world decisions. For further definitions, please refer to Appendix~\ref{appendix_background}.

\subsection{Types of Time Series Reasoning}
\label{section_3_2}
Time series reasoning seeks to understand how sequence patterns evolve over time and what drives those changes. Accordingly, we categorize it along two dimensions: (1) \textbf{Reasoning Structure} and (2) \textbf{Task Objective}.

\paragraph{Reasoning Structure.} 
A reasoning structure defines how observations lead to conclusions~\cite{wei2022chain, chu2023survey, zhang2025system}. We identify four types. (1) \textit{End-to-end} reasoning maps inputs directly to outputs with implicit reasoning. (2) \textit{Forward} reasoning derives conclusions step by step from observed data along a linear chain. (3) \textit{Backward} reasoning starts from a target outcome or hypothesis and traces back to supporting evidence. (4) \textit{Forward-backward} reasoning alternates between proposing candidate answers and verifying them with evidence, potentially repeating until convergence. These reasoning structures can also be represented with \textit{chain}-based, \textit{tree}-based, or \textit{graph}-based formalisms to capture linear, branching, or general dependency relationships~\cite{besta2025demystifying}. Detailed descriptions are provided in Appendix~\ref{appendix_reasoning_structure}.
\vspace{-3mm}
\paragraph{Task Objective.} The analytic goal determines which reasoning approach is most appropriate. For anomaly explanation or causal analysis, etiological reasoning helps pinpoint the sources of abrupt changes. For forecasting or validating periodicity hypotheses, deductive reasoning is often more appropriate. Accordingly, the task objective guides both the selection of the reasoning type and how it is applied. Table~\ref{tab:additional_reasoning_types} summarizes reasoning types and application examples.

\section{Gap \#1: Model Design}
\label{section_3}
\begin{figure*}[!t]
\begin{center}
\includegraphics[width = \linewidth]{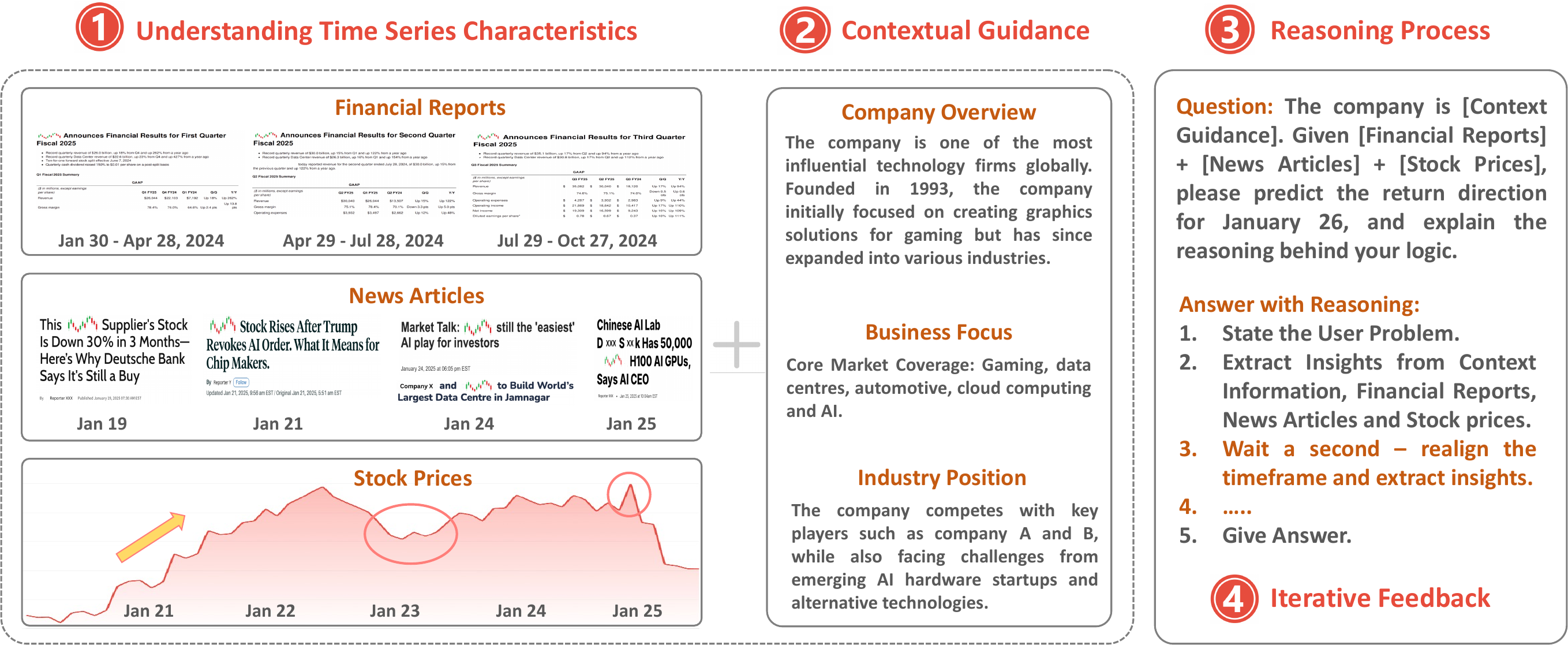}
\end{center}
\vspace{-2mm}
\caption{Key components for achieving time series reasoning (illustrated with a financial time series example).}
\vspace{-3mm}
\label{Figure_3_3}
\end{figure*}

Robust time series reasoning demands understanding temporal patterns alongside the ability to align and verify relevant context. In practice, models often fail by overlooking these characteristics, relying on static or misaligned context, or lacking mechanisms for self-checking. Accordingly, we propose four essential components for effective time series reasoning: (1) \textit{Understanding Time Series Characteristics}, (2) \textit{Contextual Guidance}, (3) \textit{Reasoning Process}, and (4) \textit{Iterative Feedback} (illustrated in Figure \ref{Figure_3_3}).

\subsection{Understanding Time Series Characteristics} 
\label{sec3.1}
\paragraph{Key Observations.}
A fundamental challenge in time series reasoning is effectively representing continuous numerical data for LLMs designed for discrete text. Current literature explores four approaches, each with trade-offs:
\vspace{-2mm}
\begin{enumerate} [leftmargin=*]
    \item \textbf{Raw Values.} This represents time series as sequences of numerical text tokens, such as \enquote{24.5, 25.1, ...}~\cite{kong2025time, guan2025timeomni}.  While it preserves exact values and allows for direct prompting, it often struggles with long sequences due to token limits and may fail to capture high-frequency patterns~\cite{gruver2024large}.
    \item \textbf{Tokenization.} Methods such as ChatTime \cite{wang2025chattime}, TokenCast \cite{tao2025values}, and Chronos \cite{ansari2024chronos} map continuous data into discrete symbols via structured delimiters (e.g., \#\#\#), vector quantization, or fixed-vocabulary binning. This makes temporal patterns more explicit and easier to learn. However, this inherently reduces precision and complicates alignment across multivariate channels with different scales.
    \item \textbf{Encoding.} This uses specialized encoders such as patch-based (Time-LLM~\cite{jin2024time}), learnable soft prompt (OpenTSLM~\cite{langer2025opentslm}) or frequency-domain encoders (FreqLLM~\cite{10.24963/ijcai.2025/377}) to compress time series into dense embeddings before feeding them to the LLM. This is efficient for long sequences but can create a \enquote{black box}, making the reasoning process less interpretable compared to text-based methods. 
    \item \textbf{Cross-Modality Transfer.} Recent works such as MLLM4TS~\cite{liu2025mllm4ts} and VisionTS++~\cite{shen2025visionts++} convert time series into visual plots and leverage vision-LLM to \enquote{see} patterns. This effectively captures global trends and handles long histories but may lose precision in fine-grained numerical reasoning. 
\end{enumerate}

\paragraph{\textcolor{blue}{Critical Gap.}}
The question remains which representation is optimal. Tokenization offers interpretability and alignment with language tasks, but it suffers from redundancy. Encoders capture complex dependencies but disrupt the flow of language processing. Visual approaches excel at macroscopic patterns but struggle with microscopic precision. The lack of consensus reveals a gap: \textit{We lack a unified representation that simultaneously ensures interpretability, computational efficiency, and precise temporal alignment.}

\paragraph{Empirical Insights with Open Questions.}
Despite the proliferation of methods, key questions remain: \textit{Under what circumstances does each representation fail? How can we select the most appropriate method for a given dataset?} Our empirical analysis suggests the following (please see Appendix~\ref{appendix_time_series_characteristics} for experiment details and figure illustration):

\begin{itemize} [leftmargin=*]
    \item \textbf{Raw values struggle with length.} Feeding raw numerical sequences directly into LLMs degrades rapidly as length increases, likely due to the \enquote{lost-in-the-middle} phenomenon and excessive token usage.
    \item \textbf{Multivariate adds complexity.} While discretization can stabilize univariate reasoning, multivariate series introduce additional failure modes: cross-channel misalignment, vocabulary explosion, and ambiguity in whether tokens represent within- or cross-channel structure.
    \item \textbf{Encoding sacrifices interpretability.} Encoder-based methods often lack verifiable explanations, as the LLM reasons over embeddings rather than human-readable values, making it hard to trace predictions to specific inputs.
    \item \textbf{Visual modality helps long-context reasoning.} Converting time series to images improves long-horizon performance by allowing models to process trends and patterns holistically, bypassing token-length constraints.
\end{itemize}

\subsection{Contextual Guidance} 
\paragraph{Key Observations.}
Context significantly influences time series interpretation, as the same data can yield vastly different predictions depending on accompanying information \cite{hu2025contextalignment, lee2025timecap}. Context may be internal (seasonality or trends) or external (economic indicators or news). Current literature explores four approaches:
\vspace{-2mm}
\begin{enumerate} [leftmargin=*]
    \item \textbf{Static.} Fixed prompts or instructions such as \enquote{this time series represents patient heart rate measurements...} incorporate domain knowledge and statistical information about the time series~\cite{lee2025map4ts, hu2025contextalignment}.
    \item \textbf{Few-Shot Learning.} Providing example time series windows~\cite{gopali2025context} or QA pairs~\cite{parker2025augmenting, tang2025electro} helps guide the model's predictions and reasoning style, though such examples are typically static and may not adapt to temporal variations.
    \item \textbf{Retrieval-Augmented (RAG).} Information such as news, domain knowledge, or similar time series is retrieved from pre-collected knowledge bases~\cite{yang2025timerag, ning2025tsrag} or searched in real-time via agentic way~\cite{zhao2025timeseriesscientist, li2026findeepforecast}.
    \item \textbf{Cross-Modal.} Aligned images are integrated as context, allowing models to leverage visual information for richer understanding~\cite{liu2025mllm4ts, siru2025time}. Video and audio modalities remain largely underexplored.
\end{enumerate}
\vspace{-5mm}
\paragraph{\textcolor{blue}{Critical Gap.}}
Current approaches rely on static inputs such as fixed prompts and pre-collected knowledge bases, which offer consistency but struggle when data distributions shift. Agentic systems for dynamic retrieval offer an alternative, yet whether they introduce more noise than signal remains unresolved. Verification mechanisms~\cite{strong2025tsver} to ensure retrieved context is accurate, relevant, and temporally aligned are still lacking. This highlights a critical gap: \textit{We lack principled understanding of how contextual guidance improves reasoning and how to integrate dynamic context without false correlations.}
\vspace{-2mm}
\paragraph{Empirical Insights with Open Questions.} 
\textit{How should context be integrated through static prompts, few-shot learning, dynamic retrieval, or cross-modal fusion to enhance the performance?} Our empirical findings reveal (please see Appendix~\ref{appendix_contextual_guidance} for experiment details and figure illustration):
\vspace{-2mm}
\begin{itemize} [leftmargin=*]
    \item \textbf{Static context has limitations.} Background-only prompts are often too broad and introduce noise that degrades quality when irrelevant context is incorporated.
    \item \textbf{Few-shot learning struggles with temporal adaptation.} Static examples lose effectiveness when time series characteristics evolve and fail to reflect current patterns or distribution shifts. Adaptive example selection that accounts for temporal dynamics remains an open challenge.
    \item \textbf{Dynamic retrieval shows promise but lacks verification.} Web-based retrieval improves relevance but often suffers from temporal misalignment between the retrieved context and the analyzed data. Verification mechanisms are essential to filter outdated or irrelevant information.
    \item \textbf{Cross-modal methods remain underexplored.} Alignment between time series and visual/audio remains challenging, with limited understanding of how to ensure temporal correspondence and avoid information leakage.
\end{itemize}

\subsection{Reasoning Process}
\paragraph{Key Observations.}
Time series reasoning aims to unify multiple objectives within an integrated framework, facing three challenges: maintaining consistent logic across steps; incorporating external context without spurious correlations; and articulating inference steps when revising earlier conclusions. Current literature explores five training techniques:
\vspace{-6mm}
\begin{enumerate} [leftmargin=*]
    \item \textbf{Adaptation.} Lightweight components (e.g., adapters) align LLMs with time series tasks, optimizing for task-specific outcomes while reasoning remains implicit~\cite{jin2024time, kong2025fusing, hwang2025decision}.
    \item \textbf{Rationale SFT.} Supervised fine-tuning (SFT) incorporates explicit reasoning traces, such as CoT or rationale explanations, promoting step-by-step inference~\cite{kong2025time, yang2025time, lan2025gem}.
    \item \textbf{Two-Stage Representation-to-Reasoning.} Training proceeds in two stages: first aligning or encoding time series data, then performing SFT for task-specific reasoning, enabling better input understanding and stronger downstream reasoning~\cite{xie2025chatts}.
    \item \textbf{Preference and Reinforcement Learning (RL).} Training optimizes preferences or verifiable reward signals (e.g., GRPO-style RL) to strengthen multi-step time series reasoning beyond supervised labels~\cite{guan2025timeomni, parker2025eliciting, zhang2025timemaster}.
    \item \textbf{Agentic.} Agentic workflows, either single- or multi-agent, decompose time series reasoning into planning, tool-based analysis, forecasting, and report synthesis~\cite{yu2025ts, zhao2025timeseriesscientist}.
\end{enumerate}

\paragraph{\textcolor{blue}{Critical Gap.}}
A key unclear point is which training techniques are effective for time series reasoning. Most research directly adapts NLP techniques, but whether these methods are optimal is unknown. An emerging trend incorporates LLMs with time series foundation models, using LLMs as explainers rather than direct forecasters~\cite{cao2025conversational, 10.1145/3768292.3771251, yeh2025empowering}. While promising, fundamental questions remain: \textit{Which training techniques are most effective? When do borrowed NLP techniques succeed or fail? Do we need fundamentally different approach that explicitly encodes time series reasoning, or can we achieve effective integration through better adaptation?}
\vspace{-2mm}
\paragraph{Empirical Insights with Open Questions.} 
\textit{Are NLP fine-tuning techniques useful for time series reasoning?} Our empirical analysis suggests the following (please see Appendix~\ref{appendix_reasoning_process} for experiment details and figure illustration):
\vspace{-4mm}
\begin{itemize} [leftmargin=*]
    \item \textbf{General-purpose LLMs outperform instruction-tuned models.} ChatGPT-5.2-thinking and Qwen3-Max without time series tuning outperform tuned models, and their rapid iteration quickly surpasses tuned variants, questioning instruction tuning's cost-effectiveness.
    \item \textbf{Long time series challenge general-purpose LLMs.} As observed in Section~\ref{sec3.1}, general-purpose LLMs without time series-specific tuning fail on long time series when given raw data, revealing limitations of direct NLP adaptation. This prompts reconsideration: \textit{Should time series models handle processing while LLMs handle explanation? What is the optimal integration approach?}
\end{itemize}

\vspace{-2mm}
\subsection{Iterative Feedback}
\paragraph{Key Observations.}
Iterative feedback enables models to refine reasoning by identifying inconsistencies, revising conclusions, and incorporating new information. This is particularly crucial for time series reasoning, where temporal dependencies and multi-step inference require careful validation. Current literature explores three approaches:
\vspace{-2mm}
\begin{enumerate} [leftmargin=*]
     \item \textbf{Self-Evaluation.} The model evaluates its own outputs or intermediate reasoning steps, identifying potential errors or inconsistencies before finalizing conclusions~\cite{bazaga2025learning, zhou2025enhancing}.
     \item \textbf{Tool-Augmented Verification.} External tools, LLM-based agents, or domain experts validate reasoning steps against established knowledge, identifying inconsistencies and providing feedback for correction~\cite{su2025chain, jalori-etal-2025-flairr, strong2025tsver}.
     \item \textbf{Multi-Agent Debate.} Multiple agents engage in structured discussion, refining reasoning through iterative critique and consensus-building~\cite{zhang2025can}.
\end{enumerate}

\paragraph{\textcolor{blue}{Critical Gap.}}
While GRPO-style reinforcement learning has been widely applied to optimize reasoning outputs~\cite{guan2025timeomni}, iterative feedback at inference time remains underexplored. Existing approaches rely on LLM-as-judge agents~\cite{jalori-etal-2025-flairr}, which are prone to bias and often fail to detect time-series–specific errors such as temporal misalignment or horizon inconsistencies. This leaves a critical gap: \textit{How to design iterative feedback that improves reasoning, validates time series characteristics, and enforces faithfulness through context verification?}

\paragraph{Empirical Insights with Open Questions.} 
\textit{How effective is iterative feedback through multihop conversation for improving time series reasoning?} We conducted experiments where human provide feedbacks to Hunyuan-2.0-Think during multihop conversation, guiding models on time series characteristics, contextual alignment, and verification processes. Our analysis suggests (please see Appendix~\ref{appendix_iterative_feedback} for experiment details and figure illustration):
\vspace{-2mm}
\begin{itemize} [leftmargin=*]
    \item \textbf{Multihop conversation improves reasoning.} Models receiving human feedback demonstrate improved reasoning compared to single-pass approaches, with interactive guidance helping it identify and correct errors in real-time.
    \item \textbf{Open questions remain.} How many rounds are optimal? Which feedback forms (hints or corrections) best improve temporal validity? How can we prevent overfitting to feedback assumptions? Answering these would clarify when multihop conversational feedback is practical.
\end{itemize}
\vspace{-2mm}
\begin{tcolorbox}[top=1pt, bottom=1pt, left=1pt, right=1pt]
  \textbf{Call to Action:}~\textit{We urge the community to reconsider how models are designed for time series reasoning. This entails deeper comprehension of temporal properties, seamless integration of external context, appropriate learning paradigms, and implementation of iterative feedback.}
\end{tcolorbox} 

\section{Gap \#2: Task Formulation}
\label{section_4}
\begin{figure}
\begin{center}
\includegraphics[width = 1\linewidth]{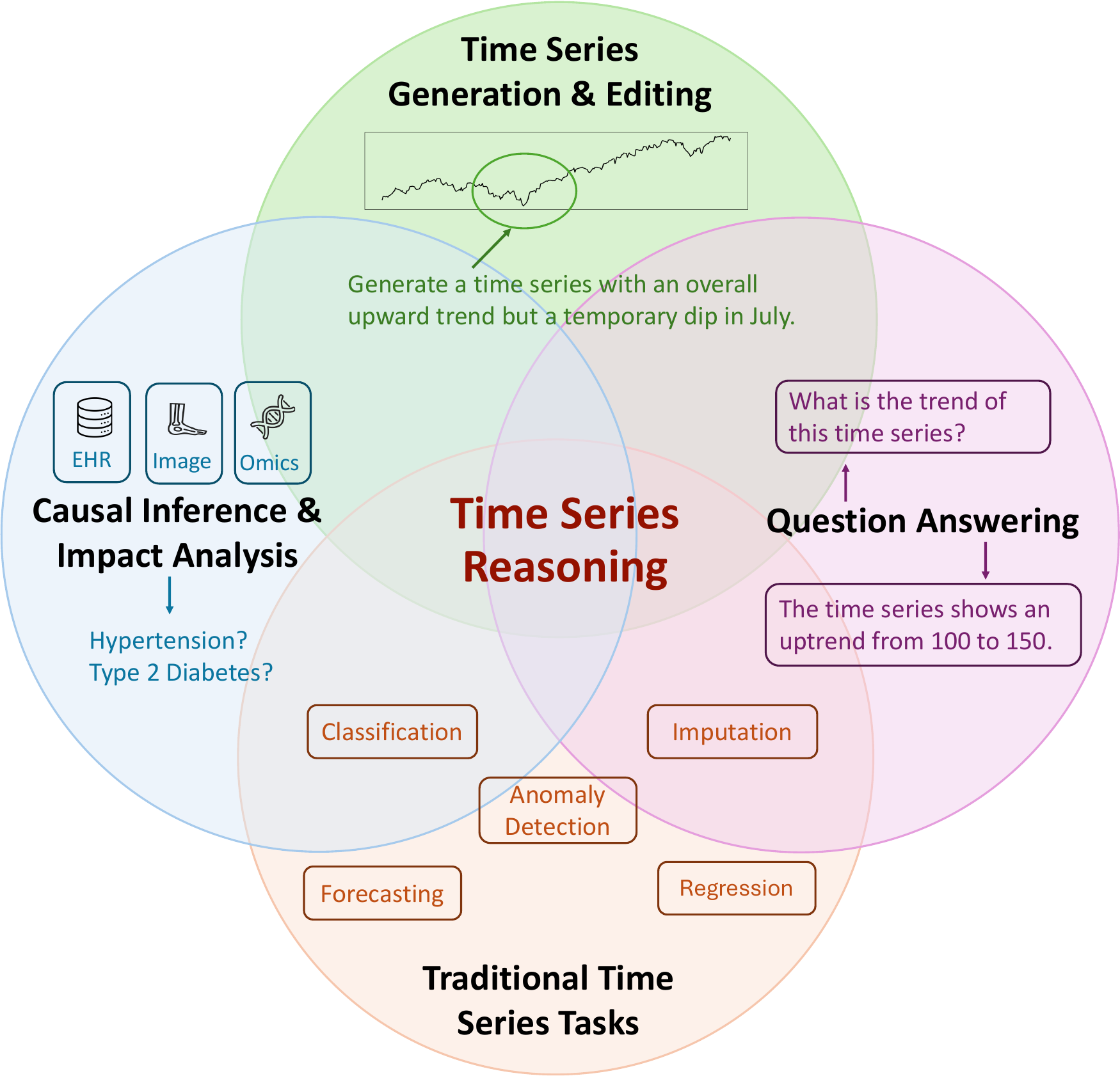}
\end{center}
\vspace{-2mm}
\caption{Illustration of diverse time series reasoning tasks.}
\vspace{-5mm}
\label{Figure_4}
\end{figure}

Time series reasoning has inspired new problem settings beyond traditional time series tasks (illustrated in Figure \ref{Figure_4}). These include \textit{question answering}, \textit{causal inference \& impact analysis}, and \textit{time series generation \& editing}, which focus on reasoning and creative manipulation of time series. This section examines these emerging tasks and reveals critical gaps in current task formulations.

\begin{table*}[!t]
\caption{Selected dataset and benchmarks for time series reasoning.}
\vspace{-2mm}
\resizebox{\linewidth}{!}{
\renewcommand{\arraystretch}{1}{
\begin{tabular}{lccccccccc}
\toprule            
\textbf{Dataset} & \textbf{Time Series} & \textbf{Text} & \textbf{Image}  & \textbf{URL} & \textbf{Application Domain} & \textbf{Task Type} & \textbf{Algorithm} & \textbf{Resource} \\ \midrule
TimeSerieseExam~\cite{cai2024timeseriesexam} &  \ding{52} &  \ding{56} & \ding{56} & \ding{56} & General & Q\&A & MLLM-based & \href{https://huggingface.co/datasets/AutonLab/TimeSeriesExam1}{[Link]} \\ \midrule
TimeMMD~\cite{liutime}  &  \ding{52} &  \ding{52} & \ding{56} & \ding{52} & General & Forecasting & MLLM-based & \href{https://github.com/AdityaLab/Time-MMD}{[Link]} \\ \midrule
ChatTime~\cite{wang2025chattime}  &  \ding{52} &  \ding{52} & \ding{56} & \ding{56} & General & Q\&A & LLM-based & \href{https://huggingface.co/datasets/ChengsenWang/TSQA}{[Link]} \\ \midrule
ChatTS~\cite{xie2025chatts}  &  \ding{52} &  \ding{52} & \ding{56} & \ding{56} & General & Q\&A & MLLM-based & \href{https://huggingface.co/ChatTSRepo}{[Link]} \\ \midrule
CiK~\cite{williams2024context}  &  \ding{52} &  \ding{52} & \ding{56} & \ding{56} & General & Forecasting & LLM-based & \href{https://huggingface.co/datasets/ServiceNow/context-is-key}{[Link]} \\ \midrule
TSQA~\cite{kong2025time}  &  \ding{52} &  \ding{52} & \ding{56} & \ding{52} & General & 
Q\&A & LLM-based & \href{https://huggingface.co/datasets/Time-MQA/TSQA}{[Link]} \\ \midrule
EngineMT-QA~\cite{wang2025itformer} &  \ding{52} &  \ding{52} & \ding{56} & \ding{56} & Engineering & Q\&A & LLM-based & \href{https://github.com/Pandalin98/ITFormer-ICML25}{[Link]} \\ \midrule
DeepFund~\cite{li2025timetravel} &  \ding{52} &  \ding{52} & \ding{56} & \ding{52} & Finance & 
Decision-Making & LLM-based & \href{https://github.com/HKUSTDial/DeepFund}{[Link]} \\ \midrule
RATs40K~\cite{yang2025time} &  \ding{52} &  \ding{52} & \ding{52} & \ding{52} & General & 
Anomaly Reasoning & MLLM-based & \href{https://huggingface.co/datasets/Time-RA/RATs40K}{[Link]} \\ \midrule
ECG-Grounding~\cite{lan2025gem} &  \ding{52} &  \ding{52} & \ding{52} & \ding{56} & Healthcare & 
Q\&A & MLLM-based & \href{https://huggingface.co/datasets/LANSG/ECG-Grounding}{[Link]} \\
\bottomrule
\end{tabular}}}
\vspace{-4mm}
\label{tab:dataset}
\end{table*}

\subsection{Question Answering}
\paragraph{Key Observations.} 
Time series-based QA has emerged as a distinct research direction, shifting from classical analysis to high-level reasoning~\cite{su2025chain, liu2026rationale, wang2025itformer, kong2025time}. These tasks take time series as primary inputs, optionally enriched with multimodal data for context, with formats ranging from open-ended to multiple choice and true/false questions. Applications span healthcare~\cite{langer2025opentslm}, meteorology~\cite{he2025radarqa}, and industrial systems~\cite{park2026bridging}.
\vspace{-3mm}
\paragraph{Empirical Insights.} 
We analyze QA reasoning using different MLLMs on healthcare data (Figure \ref{fig:case_healthcare}, Appendix \ref{appendix_question_answering}). Given ECG recordings and background information, MLLMs are tasked with detecting anomalies, identifying illnesses, and providing diagnostic reasoning based on temporal pattern analysis and medical knowledge.
\vspace{-2mm}
\subsection{Causal Inference and Impact Analysis}
\paragraph{Key Observations.} 
Time series causal inference and impact analysis uncover causal relationships, quantify event effects, and enable counterfactual reasoning~\cite{cui2025augur, chukwu2025counterfactual, rodling2025causal}. Applications span finance, where stock prices combined with financial reports reveal market impacts~\cite{zhang2025camef}, healthcare, where integrated health records evaluate intervention effects~\cite{zhang2025medkgent}, and marketing and public policy~\cite{cerutti2025methodological, niu2025event}. Additionally, knowledge graphs can be integrated to ground causal relationships and further enhance reasoning~\cite{sun2025timemkg}.
\vspace{-2mm}
\paragraph{Empirical Insights.} 
We analyze causal reasoning using different MLLMs on financial data with and without reports (Figure \ref{fig:case_financial}, Appendix \ref{appendix_casual_inference}). Given stock price sequences and background information, MLLMs are tasked with detecting anomalies, providing causal inference about price fluctuations, and analyzing market prospects based on temporal pattern analysis and financial knowledge.

\subsection{Time Series Generation and Editing}
\paragraph{Key Observations.} 
Time series generation and editing synthesize or modify time series, optionally enriched by multimodal data~\cite{pmlr-v267-li25ah}. Applications include energy systems, where synthetic electricity consumption data simulates rare scenarios~\cite{fuest2025cents}, and scientific domains requiring controllable temporal synthesis~\cite{wu2025scits}. Multimodal inputs such as satellite imagery and demographic data ensure alignment with temporal dependencies while enhancing realism~\cite{siru2025time}.
\vspace{-2mm}
\paragraph{Empirical Insights.} 
We test MLLMs on household electricity consumption with missing values (Figure \ref{fig:case_electricity}, Appendix \ref{appendix_time_series_generation}). The task requires imputing gaps and generating complete sequences by reasoning about temporal dependencies, including daily usage cycles and seasonal variations.
\vspace{-5mm}
\paragraph{\textcolor{blue}{Critical Gap.}}
Current research predominantly focuses on templated QA with multiple-choice or true/false formats. While incorporating causal analysis, they remain confined to the QA format, potentially since evaluation beyond accuracy is challenging. However, real-world problems are not multiple-choice questions. Consider the following example:
\begin{tcolorbox}[colback=gray!5!white,colframe=gray!75!black,title=Example QA Task]
\textbf{Task:} You are given a time series. Please identify the scenario that most likely created it? \
\textbf{Input:} [2879, 3858, 4301, 4380,...]\
\textbf{Options:} (A) A mall launches a limited-time promotion...; (B) A local bakery records...; (C) A regional soccer championship...; (D) Monthly sales data of a toy store...
\end{tcolorbox}
Such tasks, while useful for benchmarking, bear little resemblance to real-world needs where practitioners face open-ended questions without ground truth. Moreover, while LLMs excel at identifying trends, direct forecasting remains unreliable. However, evaluating open-ended responses remains challenging, which we discuss in Gap \#3. The gap is clear: \textit{current task formulations are either simplistic or fail to reflect real-world decision-making scenarios.}

\begin{tcolorbox}[top=1pt, bottom=1pt, left=1pt, right=1pt]
  \textbf{Call to Action:}~\textit{Current task formulations constrain the potential of time series reasoning. To address this, researchers should drive progress on open-ended QA, causal discovery, and scenario analysis that translate benchmark performance into actionable insights.}
\end{tcolorbox}

\section{Gap \#3: Evaluation}
\label{section_5}
Evaluating time series reasoning faces fundamental challenges: reasoning quality is difficult to quantify objectively, and existing benchmarks fail to fully capture real-world complexity. This section examines existing datasets and metrics, identifies critical gaps, and propose improvement.
\subsection{Dataset and Benchmark} 
\paragraph{Key Observations.} 
A substantial number of datasets have been released since 2025, indicating growing research interest. We present representative examples in Table~\ref{tab:dataset}.
\vspace{-2mm}
\paragraph{Empirical Insights with Open Questions.}
Current benchmarks are typically QA-based, constructed by partitioning real datasets and appending task instructions with textual context and reasoning steps. This raises an important question: \textit{beyond QA formats, could future research design scenario-based benchmarks grounded in real-world decision-making tasks?} Developing such benchmarks would require collaboration with domain experts. The case study in our introduction illustrates one such example.

\subsection{Evaluation Metrics} 
\paragraph{Key Observations.} 
Reasoning is inherently subjective and difficult to evaluate objectively. Current practices primarily rely on: (1) \textit{expert-annotated user studies} assessing reasoning quality; and (2) \textit{benchmark-based assessments} measuring accuracy for QA or MSE/MAE for forecasting. However, these methods rarely test scenarios requiring multi-step reasoning, evidence verification, or handling of novel situations such as conflicting information.
\vspace{-2mm}
\paragraph{Empirical Insights with Open Questions.}
User studies are complicated by the fact that different experts often have divergent views and preferences. Moreover, we observe that different LLMs produce different reasoning paths for the same question, and even the same LLM generates varying outputs across runs~\cite{chang2024survey}. This raises a key question: \textit{how can evaluation frameworks account for this variability while ensuring fairness?}
\vspace{-2mm}
\paragraph{\textcolor{blue}{Critical Gap.}}
Current evaluation faces several challenges. We identify key gaps and propose a protocol in Appendix~\ref{appendix_evaluation}:
\vspace{-6mm}
\begin{itemize} [leftmargin=*]
    \item \textbf{Benchmark realism.} Many benchmarks reduce real-world time series problems to static QA formats, missing streaming dynamics, decision loops, and operational constraints. Scenario-based assessments remain unexplored.
    \item \textbf{Limited multimodality.} Benchmarks often omit other modalities commonly available in practice (event, documents, images, or sensor fusion), limiting general applicability and masking failure modes in multimodal reasoning.
    \item \textbf{Narrow reasoning coverage.} Benchmarks emphasize forward reasoning (e.g., chain-of-thought), leaving gaps in diverse reasoning structures discussed in Section~\ref{section_2}.
    \item \textbf{User study inconsistency.} Divergent expert preferences and LLM stochasticity complicate fair evaluation. We envision standardized frameworks are needed for consistency through aligned rubrics and multi-view judgment.
\end{itemize}

\begin{tcolorbox}[top=1pt, bottom=1pt, left=1pt, right=1pt]
  \textbf{Call to Action:}~\textit{Realizing the full potential of time series reasoning requires rethinking how performance is measured. Future evaluation efforts should emphasize practical benchmarks, multimodal integration, reasoning structures, and robust protocols for expert assessment.}
\end{tcolorbox}

\section{Alternative Views}
\paragraph{\textit{Is Time Series Reasoning Necessary, or Are Traditional Time Series Tasks Sufficient?}}
One might argue that traditional time series tasks are sufficient. Tasks such as forecasting, classification, and anomaly detection perform well when objectives are clearly specified and the relevant signals are contained within the observed data. If the goal is primarily predictive accuracy under relatively stable conditions, these tasks can be more efficient and reliable than open-ended reasoning-based approaches.

However, many real-world uses of time series go beyond prediction. Decision makers often need to understand \textit{why} a pattern occurred, assess competing explanations, and determine \textit{what} action to take under uncertainty. As illustrated in our introductory case study, analysts need to integrate contextual guidance beyond numerical trajectories (\textbf{Gap 1}, Section~\ref{section_3}), reason about interventions, causality, and decision-making under uncertainty (\textbf{Gap 2}, Section~\ref{section_4}), and provide explanations that are faithful and interpretable, properties that accuracy-centric evaluations do not adequately measure (\textbf{Gap 3}, Section~\ref{section_5}). Time series reasoning directly targets these requirements by supporting explicit inference, context integration, and responses to open-ended, decision-oriented queries. In this sense, traditional tasks remain essential baselines for well-structured prediction problems, but reasoning capabilities are necessary when explanation, robustness, and actionability are first-class objectives.
\vspace{-2mm}
\paragraph{\textit{Should We Pursue Time Series Reasoning with LLMs/MLLMs, or Are Specialized Time Series Models More Appropriate?}}
One might argue that specialized time series models are more appropriate. They excel at temporal pattern recognition and numerical prediction, are computationally efficient, and have been extensively validated. From this perspective, research should prioritize strengthening these models rather than adapting LLMs/MLLMs that may not be suited to temporal dynamics or numerical precision.

However, our analysis suggests that the choice is not purely either or. LLM-based approaches face obstacles including difficulty handling long sequences and sensitivity to representation choices (\textbf{Gap 1}, Section~\ref{section_3}), while specialized models excel at pattern matching but lack mechanisms for explanation, context integration, or open-ended questions. As \textbf{Gap 2} highlights, real-world applications require causal inference and decision-making under uncertainty (Section~\ref{section_4}). Therefore, the more promising direction could be hybrid systems that combine specialized models for numerical precision and temporal structure with LLMs for reasoning, explanation, and multimodal or contextual integration. Consistent with our argument across all three gaps, progress will require unified frameworks that jointly align model design, task formulation, and evaluation.

\section{Conclusion}
Time series analysis is moving beyond prediction toward interactive, context-aware understanding, driven by growing demand for decision support and heterogeneous data. \textbf{We argue that achieving time series reasoning requires fundamentally rethinking model design, task formulation, and evaluation together}. We call on researchers to pursue this shift: models should better represent temporal structure and enforce faithfulness; tasks should reflect practice through causal analysis and scenario-driven generation; and evaluation should move beyond benchmark accuracy toward reasoning quality, robustness, and decision relevance. Treating benchmarks as diagnostic tools rather than endpoints will advance research toward time series intelligence.

\bibliography{reference}

@article{moor2023foundation,
  title={Foundation models for generalist medical artificial intelligence},
  author={Moor, Michael and Banerjee, Oishi and Abad, Zahra Shakeri Hossein and Krumholz, Harlan M and Leskovec, Jure and Topol, Eric J and Rajpurkar, Pranav},
  journal={Nature},
  volume={616},
  number={7956},
  pages={259--265},
  year={2023},
  publisher={Nature Publishing Group UK London}
}

@article{hollmann2025accurate,
  title={Accurate predictions on small data with a tabular foundation model},
  author={Hollmann, Noah and M{\"u}ller, Samuel and Purucker, Lennart and Krishnakumar, Arjun and K{\"o}rfer, Max and Hoo, Shi Bin and Schirrmeister, Robin Tibor and Hutter, Frank},
  journal={Nature},
  volume={637},
  number={8045},
  pages={319--326},
  year={2025},
  publisher={Nature Publishing Group UK London}
}

@inproceedings{jin2024time,
  title={Time-LLM: Time Series Forecasting by Reprogramming Large Language Models},
  author={Jin, Ming and Wang, Shiyu and Ma, Lintao and Chu, Zhixuan and Zhang, James Y and Shi, Xiaoming and Chen, Pin-Yu and Liang, Yuxuan and Li, Yuan-Fang and Pan, Shirui and Wen, Qingsong},
  booktitle={International Conference on Learning Representations (ICLR)},
  year={2024}
}

@inproceedings{liutime,
  title={{Time-MMD}: Multi-Domain Multimodal Dataset for Time Series Analysis},
  author={Liu, Haoxin and Xu, Shangqing and Zhao, Zhiyuan and Kong, Lingkai and Kamarthi, Harshavardhan and Sasanur, Aditya B and Sharma, Megha and Cui, Jiaming and Wen, Qingsong and Zhang, Chao and others},
  booktitle={The Thirty-eight Conference on Neural Information Processing Systems Datasets and Benchmarks Track},
  year={2024}

}

@article{yang2024survey,
  title={A survey on diffusion models for time series and spatio-temporal data},
  author={Yang, Yiyuan and Jin, Ming and Wen, Haomin and Zhang, Chaoli and Liang, Yuxuan and Ma, Lintao and Wang, Yi and Liu, Chenghao and Yang, Bin and Xu, Zenglin and others},
  journal={arXiv preprint arXiv:2404.18886},
  year={2024}
}

@inproceedings{wu2023multimodal,
  title={Multimodal large language models: A survey},
  author={Wu, Jiayang and Gan, Wensheng and Chen, Zefeng and Wan, Shicheng and Philip, S Yu},
  booktitle={2023 IEEE International Conference on Big Data (BigData)},
  pages={2247--2256},
  year={2023},
  organization={IEEE}
}

@article{wen2022transformers,
  title={Transformers in time series: A survey},
  author={Wen, Qingsong and Zhou, Tian and Zhang, Chaoli and Chen, Weiqi and Ma, Ziqing and Yan, Junchi and Sun, Liang},
  journal={arXiv preprint arXiv:2202.07125},
  year={2022}
}

@article{zamanzadeh2024deep,
  title={Deep learning for time series anomaly detection: A survey},
  author={Zamanzadeh Darban, Zahra and Webb, Geoffrey I and Pan, Shirui and Aggarwal, Charu and Salehi, Mahsa},
  journal={ACM Computing Surveys},
  volume={57},
  number={1},
  pages={1--42},
  year={2024},
  publisher={ACM New York, NY}
}

@article{yin2023survey,
  title={A survey on multimodal large language models},
  author={Yin, Shukang and Fu, Chaoyou and Zhao, Sirui and Li, Ke and Sun, Xing and Xu, Tong and Chen, Enhong},
  journal={arXiv preprint arXiv:2306.13549},
  year={2023}
}

@article{zhang2024mm,
  title={Mm-llms: Recent advances in multimodal large language models},
  author={Zhang, Duzhen and Yu, Yahan and Dong, Jiahua and Li, Chenxing and Su, Dan and Chu, Chenhui and Yu, Dong},
  journal={arXiv preprint arXiv:2401.13601},
  year={2024}
}

@inproceedings{cui2024survey,
  title={A survey on multimodal large language models for autonomous driving},
  author={Cui, Can and Ma, Yunsheng and Cao, Xu and Ye, Wenqian and Zhou, Yang and Liang, Kaizhao and Chen, Jintai and Lu, Juanwu and Yang, Zichong and Liao, Kuei-Da and others},
  booktitle={Proceedings of the IEEE/CVF Winter Conference on Applications of Computer Vision},
  pages={958--979},
  year={2024}
}

@article{du2024tsi,
  title={Tsi-bench: Benchmarking time series imputation},
  author={Du, Wenjie and Wang, Jun and Qian, Linglong and Yang, Yiyuan and Ibrahim, Zina and Liu, Fanxing and Wang, Zepu and Liu, Haoxin and Zhao, Zhiyuan and Zhou, Yingjie and others},
  journal={arXiv preprint arXiv:2406.12747},
  year={2024}
}

@article{zhou2024survey,
  title={A survey on generative ai and llm for video generation, understanding, and streaming},
  author={Zhou, Pengyuan and Wang, Lin and Liu, Zhi and Hao, Yanbin and Hui, Pan and Tarkoma, Sasu and Kangasharju, Jussi},
  journal={arXiv preprint arXiv:2404.16038},
  year={2024}
}

@article{mohammadi2024deep,
  title={Deep learning for time series classification and extrinsic regression: A current survey},
  author={Mohammadi Foumani, Navid and Miller, Lynn and Tan, Chang Wei and Webb, Geoffrey I and Forestier, Germain and Salehi, Mahsa},
  journal={ACM Computing Surveys},
  volume={56},
  number={9},
  pages={1--45},
  year={2024},
  publisher={ACM New York, NY}
}

@article{yu2024natural,
  title={Natural language reasoning, a survey},
  author={Yu, Fei and Zhang, Hongbo and Tiwari, Prayag and Wang, Benyou},
  journal={ACM Computing Surveys},
  volume={56},
  number={12},
  pages={1--39},
  year={2024},
  publisher={ACM New York, NY}
}

@article{chow2024towards,
  title={Towards time series reasoning with llms},
  author={Chow, Winnie and Gardiner, Lauren and Hallgr{\'\i}msson, Haraldur T and Xu, Maxwell A and Ren, Shirley You},
  journal={arXiv preprint arXiv:2409.11376},
  year={2024}
}

@article{chang2024survey,
  title={A survey on evaluation of large language models},
  author={Chang, Yupeng and Wang, Xu and Wang, Jindong and Wu, Yuan and Yang, Linyi and Zhu, Kaijie and Chen, Hao and Yi, Xiaoyuan and Wang, Cunxiang and Wang, Yidong and others},
  journal={ACM Transactions on Intelligent Systems and Technology},
  volume={15},
  number={3},
  pages={1--45},
  year={2024},
  publisher={ACM New York, NY}
}

@inproceedings{xie2025chatts,
  title={ChatTS: Aligning Time Series with LLMs via Synthetic Data for Enhanced Understanding and Reasoning},
  author={Xie, Zhe and Li, Zeyan and He, Xiao and Xu, Longlong and Wen, Xidao and Zhang, Tieying and Chen, Jianjun and Shi, Rui and Pei, Dan},
  booktitle={Proceedings of the VLDB Endowment, 2025},
  year={2025}
}

@article{xia2024beyond,
  title={Beyond chain-of-thought: A survey of chain-of-x paradigms for llms},
  author={Xia, Yu and Wang, Rui and Liu, Xu and Li, Mingyan and Yu, Tong and Chen, Xiang and McAuley, Julian and Li, Shuai},
  journal={arXiv preprint arXiv:2404.15676},
  year={2024}
}

@article{shi2024language,
  title={Language models can improve event prediction by few-shot abductive reasoning},
  author={Shi, Xiaoming and Xue, Siqiao and Wang, Kangrui and Zhou, Fan and Zhang, James and Zhou, Jun and Tan, Chenhao and Mei, Hongyuan},
  journal={Advances in Neural Information Processing Systems},
  volume={36},
  year={2024}
}

@article{lewis2024using,
  title={Using counterfactual tasks to evaluate the generality of analogical reasoning in large language models},
  author={Lewis, Martha and Mitchell, Melanie},
  journal={arXiv preprint arXiv:2402.08955},
  year={2024}
}

@inproceedings{yang2023dcdetector,
  title={Dcdetector: Dual attention contrastive representation learning for time series anomaly detection},
  author={Yang, Yiyuan and Zhang, Chaoli and Zhou, Tian and Wen, Qingsong and Sun, Liang},
  booktitle={Proceedings of the 29th ACM SIGKDD Conference on Knowledge Discovery and Data Mining},
  pages={3033--3045},
  year={2023}
}

@article{nie2022time,
  title={A time series is worth 64 words: Long-term forecasting with transformers},
  author={Nie, Yuqi and Nguyen, Nam H and Sinthong, Phanwadee and Kalagnanam, Jayant},
  journal={arXiv preprint arXiv:2211.14730},
  year={2022}
}

@inproceedings{jin2024position,
  title={Position: What Can Large Language Models Tell Us about Time Series Analysis},
  author={Jin, Ming and Zhang, Yifan and Chen, Wei and Zhang, Kexin and Liang, Yuxuan and Yang, Bin and Wang, Jindong and Pan, Shirui and Wen, Qingsong},
  booktitle={Forty-first International Conference on Machine Learning},
  year={2024}
}

@book{hamilton2020time,
  title={Time series analysis},
  author={Hamilton, James D},
  year={2020},
  publisher={Princeton university press}
}

@book{kirchgassner2012introduction,
  title={Introduction to modern time series analysis},
  author={Kirchg{\"a}ssner, Gebhard and Wolters, J{\"u}rgen and Hassler, Uwe},
  year={2012},
  publisher={Springer Science \& Business Media}
}

@book{shumway2000time,
  title={Time series analysis and its applications},
  author={Shumway, Robert H and Stoffer, David S and Stoffer, David S},
  volume={3},
  year={2000},
  publisher={Springer}
}

@article{besta2025reasoning,
  title={Reasoning Language Models: A Blueprint},
  author={Besta, Maciej and Barth, Julia and Schreiber, Eric and Kubicek, Ales and Catarino, Afonso and Gerstenberger, Robert and Nyczyk, Piotr and Iff, Patrick and Li, Yueling and Houliston, Sam and others},
  journal={arXiv preprint arXiv:2501.11223},
  year={2025}
}

@article{wei2022chain,
  title={Chain-of-thought prompting elicits reasoning in large language models},
  author={Wei, Jason and Wang, Xuezhi and Schuurmans, Dale and Bosma, Maarten and Xia, Fei and Chi, Ed and Le, Quoc V and Zhou, Denny and others},
  journal={Advances in neural information processing systems},
  volume={35},
  pages={24824--24837},
  year={2022}
}

@article{chu2023survey,
  title={A survey of chain of thought reasoning: Advances, frontiers and future},
  author={Chu, Zheng and Chen, Jingchang and Chen, Qianglong and Yu, Weijiang and He, Tao and Wang, Haotian and Peng, Weihua and Liu, Ming and Qin, Bing and Liu, Ting},
  journal={arXiv preprint arXiv:2309.15402},
  year={2023}
}

@article{zhou2023one,
  title={One fits all: Power general time series analysis by pretrained lm},
  author={Zhou, Tian and Niu, Peisong and Sun, Liang and Jin, Rong and others},
  journal={Advances in neural information processing systems},
  volume={36},
  pages={43322--43355},
  year={2023}
}

@article{nie2024survey,
  title={A Survey of Large Language Models for Financial Applications: Progress, Prospects and Challenges},
  author={Nie, Yuqi and Kong, Yaxuan and Dong, Xiaowen and Mulvey, John M and Poor, H Vincent and Wen, Qingsong and Zohren, Stefan},
  journal={arXiv preprint arXiv:2406.11903},
  year={2024}
}

@inproceedings{yang2023sgdp,
  title={SGDP: A stream-graph neural network based data prefetcher},
  author={Yang, Yiyuan and Li, Rongshang and Shi, Qiquan and Li, Xijun and Hu, Gang and Li, Xing and Yuan, Mingxuan},
  booktitle={2023 International Joint Conference on Neural Networks (IJCNN)},
  pages={1--8},
  year={2023},
  organization={IEEE}
}

@article{yang2021long,
  title={Long-distance pipeline safety early warning: a distributed optical fiber sensing semi-supervised learning method},
  author={Yang, Yiyuan and Zhang, Haifeng and Li, Yi},
  journal={IEEE sensors journal},
  volume={21},
  number={17},
  pages={19453--19461},
  year={2021},
  publisher={IEEE}
}

@inproceedings{yang2021early,
  title={Early safety warnings for long-distance pipelines: A distributed optical fiber sensor machine learning approach},
  author={Yang, Yiyuan and Li, Yi and Zhang, Taojia and Zhou, Yan and Zhang, Haifeng},
  booktitle={Proceedings of the AAAI Conference on Artificial Intelligence},
  volume={35},
  number={17},
  pages={14991--14999},
  year={2021}
}

@article{gruver2024large,
  title={Large language models are zero-shot time series forecasters},
  author={Gruver, Nate and Finzi, Marc and Qiu, Shikai and Wilson, Andrew G},
  journal={Advances in Neural Information Processing Systems},
  volume={36},
  year={2024}
}

@inproceedings{liang2024foundation,
  title={Foundation models for time series analysis: A tutorial and survey},
  author={Liang, Yuxuan and Wen, Haomin and Nie, Yuqi and Jiang, Yushan and Jin, Ming and Song, Dongjin and Pan, Shirui and Wen, Qingsong},
  booktitle={Proceedings of the 30th ACM SIGKDD conference on knowledge discovery and data mining},
  pages={6555--6565},
  year={2024}
}

@article{chang2023llm4ts,
  title={Llm4ts: Two-stage fine-tuning for time-series forecasting with pre-trained llms},
  author={Chang, Ching and Peng, Wen-Chih and Chen, Tien-Fu},
  journal={arXiv preprint arXiv:2308.08469},
  year={2023}
}

@article{cao2023tempo,
  title={Tempo: Prompt-based generative pre-trained transformer for time series forecasting},
  author={Cao, Defu and Jia, Furong and Arik, Sercan O and Pfister, Tomas and Zheng, Yixiang and Ye, Wen and Liu, Yan},
  journal={arXiv preprint arXiv:2310.04948},
  year={2023}
}

@article{cai2024timeseriesexam,
  title={TimeSeriesExam: A time series understanding exam},
  author={Cai, Yifu and Choudhry, Arjun and Goswami, Mononito and Dubrawski, Artur},
  journal={arXiv preprint arXiv:2410.14752},
  year={2024}
}

@article{potosnak2024implicit,
  title={Implicit Reasoning in Deep Time Series Forecasting},
  author={Potosnak, Willa and Challu, Cristian and Goswami, Mononito and Wili{\'n}ski, Micha{\l} and {\.Z}ukowska, Nina and Dubrawski, Artur},
  journal={arXiv preprint arXiv:2409.10840},
  year={2024}
}

@article{williams2024context,
  title={Context is key: A benchmark for forecasting with essential textual information},
  author={Williams, Andrew Robert and Ashok, Arjun and Marcotte, {\'E}tienne and Zantedeschi, Valentina and Subramanian, Jithendaraa and Riachi, Roland and Requeima, James and Lacoste, Alexandre and Rish, Irina and Chapados, Nicolas and others},
  journal={arXiv preprint arXiv:2410.18959},
  year={2025}
}

@article{zhou2024can,
  title={Can {LLMs} Understand Time Series Anomalies?},
  author={Zhou, Zihao and Yu, Rose},
  journal={arXiv preprint arXiv:2410.05440},
  year={2024}
}

@inproceedings{liu2025picture,
    title = "A Picture is Worth A Thousand Numbers: Enabling {LLM}s Reason about Time Series via Visualization",
    author = "Liu, Haoxin  and
      Liu, Chenghao  and
      Prakash, B. Aditya",
    editor = "Chiruzzo, Luis  and
      Ritter, Alan  and
      Wang, Lu",
    booktitle = "Proceedings of the 2025 Conference of the Nations of the Americas Chapter of the Association for Computational Linguistics: Human Language Technologies (Volume 1: Long Papers)",
    month = apr,
    year = "2025",
    address = "Albuquerque, New Mexico",
    publisher = "Association for Computational Linguistics",
    url = "https://aclanthology.org/2025.naacl-long.383/",
    doi = "10.18653/v1/2025.naacl-long.383",
    pages = "7486--7518",
    ISBN = "979-8-89176-189-6"
}

@article{tang2024large,
  title={Are Large Language Models Useful for Time Series Data Analysis?},
  author={Tang, Francis and Ding, Ying},
  journal={arXiv preprint arXiv:2412.12219},
  year={2024}
}

@article{jiang2025multi,
  title={Multi-modal Time Series Analysis: A Tutorial and Survey},
  author={Jiang, Yushan and Ning, Kanghui and Pan, Zijie and Shen, Xuyang and Ni, Jingchao and Yu, Wenchao and Schneider, Anderson and Chen, Haifeng and Nevmyvaka, Yuriy and Song, Dongjin},
  journal={arXiv preprint arXiv:2503.13709},
  year={2025}
}

@article{liu2025can,
  title={How can time series analysis benefit from multiple modalities? a survey and outlook},
  author={Liu, Haoxin and Kamarthi, Harshavardhan and Zhao, Zhiyuan and Xu, Shangqing and Wang, Shiyu and Wen, Qingsong and Hartvigsen, Tom and Wang, Fei and Prakash, B Aditya},
  journal={arXiv preprint arXiv:2503.11835},
  year={2025}
}

@article{chang2025survey,
  title={A survey of reasoning and agentic systems in time series with large language models},
  author={Chang, Ching and Shi, Yidan and Cao, Defu and Yang, Wei and Hwang, Jeehyun and Wang, Haixin and Pang, Jiacheng and Wang, Wei and Liu, Yan and Peng, Wen-Chih and others},
  journal={arXiv preprint arXiv:2509.11575},
  year={2025}
}

@article{ning2025towards,
  title={Towards Interpretable and Trustworthy Time Series Reasoning: A BlueSky Vision},
  author={Ning, Kanghui and Pan, Zijie and Jiang, Yushan and Schneider, Anderson and Nevmyvaka, Yuriy and Song, Dongjin},
  journal={arXiv preprint arXiv:2510.16980},
  year={2025}
}

@article{besta2025demystifying,
  title={Demystifying chains, trees, and graphs of thoughts},
  author={Besta, Maciej and Memedi, Florim and Zhang, Zhenyu and Gerstenberger, Robert and Piao, Guangyuan and Blach, Nils and Nyczyk, Piotr and Copik, Marcin and Kwa{\'s}niewski, Grzegorz and M{\"u}ller, Jurgen and others},
  journal={IEEE Transactions on Pattern Analysis and Machine Intelligence},
  year={2025},
  publisher={IEEE}
}

@article{zhang2025system,
  title={From system 1 to system 2: a survey of reasoning Large Language Models},
  author={Zhang, Duzhen and Li, Zhong-Zhi and Zhang, Ming-Liang and Zhang, Jiaxin and Liu, Zengyan and Yao, Yuxuan and Xu, Haotian and Zheng, Junhao and Chen, Xiuyi and Zhang, Yingying and others},
  journal={IEEE Transactions on Pattern Analysis and Machine Intelligence},
  year={2025},
  publisher={IEEE}
}

@inproceedings{kong2025time,
    title = "Time-{MQA}: Time Series Multi-Task Question Answering with Context Enhancement",
    author = "Kong, Yaxuan  and
      Yang, Yiyuan  and
      Hwang, Yoontae  and
      Du, Wenjie  and
      Zohren, Stefan  and
      Wang, Zhangyang  and
      Jin, Ming  and
      Wen, Qingsong",
    editor = "Che, Wanxiang  and
      Nabende, Joyce  and
      Shutova, Ekaterina  and
      Pilehvar, Mohammad Taher",
    booktitle = "Proceedings of the 63rd Annual Meeting of the Association for Computational Linguistics (Volume 1: Long Papers)",
    month = jul,
    year = "2025",
    address = "Vienna, Austria",
    publisher = "Association for Computational Linguistics",
    url = "https://aclanthology.org/2025.acl-long.1437/",
    doi = "10.18653/v1/2025.acl-long.1437",
    pages = "29736--29753",
    ISBN = "979-8-89176-251-0"
}

@article{guan2025timeomni,
  title={TimeOmni-1: Incentivizing Complex Reasoning with Time Series in Large Language Models},
  author={Guan, Tong and Meng, Zijie and Li, Dianqi and Wang, Shiyu and Yang, Chao-Han Huck and Wen, Qingsong and Liu, Zuozhu and Siniscalchi, Sabato Marco and Jin, Ming and Pan, Shirui},
  journal={arXiv preprint arXiv:2509.24803},
  year={2025}
}

@inproceedings{wang2025chattime,
  title={Chattime: A unified multimodal time series foundation model bridging numerical and textual data},
  author={Wang, Chengsen and Qi, Qi and Wang, Jingyu and Sun, Haifeng and Zhuang, Zirui and Wu, Jinming and Zhang, Lei and Liao, Jianxin},
  booktitle={Proceedings of the AAAI Conference on Artificial Intelligence},
  volume={39},
  number={12},
  pages={12694--12702},
  year={2025}
}

@article{tao2025values,
  title={From values to tokens: An llm-driven framework for context-aware time series forecasting via symbolic discretization},
  author={Tao, Xiaoyu and Zhang, Shilong and Cheng, Mingyue and Wang, Daoyu and Pan, Tingyue and Pan, Bokai and Zhang, Changqing and Wang, Shijin},
  journal={arXiv preprint arXiv:2508.09191},
  year={2025}
}

@article{ansari2024chronos,
  title={Chronos: Learning the Language of Time Series},
  author={Ansari, Abdul Fatir and Stella, Lorenzo and Turkmen, Caner and
          Zhang, Xiyuan and Mercado, Pedro and Shen, Huibin and
          Shchur, Oleksandr and Rangapuram, Syama Syndar and
          Pineda Arango, Sebastian and Kapoor, Shubham and
          Zschiegner, Jasper and Maddix, Danielle C. and
          Mahoney, Michael W. and Torkkola, Kari and
          Gordon Wilson, Andrew and Bohlke-Schneider, Michael and
          Wang, Yuyang},
  journal={Transactions on Machine Learning Research},
  issn={2835-8856},
  year={2024},
  url={https://openreview.net/forum?id=gerNCVqqtR}
}

@article{langer2025opentslm,
  title={Opentslm: Time-series language models for reasoning over multivariate medical text-and time-series data},
  author={Langer, Patrick and Kaar, Thomas and Rosenblattl, Max and Xu, Maxwell A and Chow, Winnie and Maritsch, Martin and Verma, Aradhana and Han, Brian and Kim, Daniel Seung and Chubb, Henry and others},
  journal={arXiv preprint arXiv:2510.02410},
  year={2025}
}

@inproceedings{10.24963/ijcai.2025/377,
author = {Wang, Shunnan and Gao, Min and Wang, Zongwei and Bai, Yibing and Jiang, Feng and Pang, Guansong},
title = {FreqLLM: frequency-aware large language models for time series forecasting},
year = {2025},
isbn = {978-1-956792-06-5},
url = {https://doi.org/10.24963/ijcai.2025/377},
doi = {10.24963/ijcai.2025/377},
booktitle = {Proceedings of the Thirty-Fourth International Joint Conference on Artificial Intelligence},
articleno = {377},
numpages = {9},
location = {Montreal, Canada},
series = {IJCAI '25}
}

@article{liu2025mllm4ts,
  title={MLLM4TS: Leveraging Vision and Multimodal Language Models for General Time-Series Analysis},
  author={Liu, Qinghua and Heshmati, Sam and Mai, Zheda and Abraham, Zubin and Paparrizos, John and Ren, Liu},
  journal={arXiv preprint arXiv:2510.07513},
  year={2025}
}

@article{shen2025visionts++,
  title={VisionTS++: Cross-Modal Time Series Foundation Model with Continual Pre-trained Vision Backbones},
  author={Shen, Lefei and Chen, Mouxiang and Liu, Xu and Fu, Han and Ren, Xiaoxue and Sun, Jianling and Li, Zhuo and Liu, Chenghao},
  journal={arXiv preprint arXiv:2508.04379},
  year={2025}
}

@inproceedings{lee2025timecap,
  title={Timecap: Learning to contextualize, augment, and predict time series events with large language model agents},
  author={Lee, Geon and Yu, Wenchao and Shin, Kijung and Cheng, Wei and Chen, Haifeng},
  booktitle={Proceedings of the AAAI Conference on Artificial Intelligence},
  volume={39},
  number={17},
  pages={18082--18090},
  year={2025}
}

@inproceedings{hu2025contextalignment,
title={Context-Alignment: Activating and Enhancing {LLM}s Capabilities in Time Series},
author={Yuxiao Hu and Qian Li and Dongxiao Zhang and Jinyue Yan and Yuntian Chen},
booktitle={The Thirteenth International Conference on Learning Representations},
year={2025},
url={https://openreview.net/forum?id=syC2764fPc}
}

@article{lee2025map4ts,
  title={MAP4TS: A Multi-Aspect Prompting Framework for Time-Series Forecasting with Large Language Models},
  author={Lee, Suchan and Choi, Jihoon and Lee, Sohyeon and Song, Minseok and Jang, Bong-Gyu and Yu, Hwanjo and Han, Soyeon Caren},
  journal={arXiv preprint arXiv:2510.23090},
  year={2025}
}

@article{gopali2025context,
  title={In-Context and Few-Shots Learning for Forecasting Time Series Data based on Large Language Models},
  author={Gopali, Saroj and Chhetri, Bipin and Giri, Deepika and Siami-Namini, Sima and Namin, Akbar Siami},
  journal={arXiv preprint arXiv:2512.07705},
  year={2025}
}

@article{parker2025augmenting,
  title={Augmenting LLMs for General Time Series Understanding and Prediction},
  author={Parker, Felix and Chan, Nimeesha and Zhang, Chi and Ghobadi, Kimia},
  journal={arXiv preprint arXiv:2510.01111},
  year={2025}
}

@misc{tang2025electro,
      title={Electrocardiogram-Language Model for Few-Shot Question Answering with Meta Learning}, 
      author={Jialu Tang and Tong Xia and Yuan Lu and Cecilia Mascolo and Aaqib Saeed},
      year={2025},
      eprint={2410.14464},
      archivePrefix={arXiv},
      primaryClass={cs.LG},
      url={https://arxiv.org/abs/2410.14464}, 
}

@inproceedings{yang2025timerag,
  title={Timerag: Boosting llm time series forecasting via retrieval-augmented generation},
  author={Yang, Silin and Wang, Dong and Zheng, Haoqi and Jin, Ruochun},
  booktitle={ICASSP 2025-2025 IEEE International Conference on Acoustics, Speech and Signal Processing (ICASSP)},
  pages={1--5},
  year={2025},
  organization={IEEE}
}

@article{li2026findeepforecast,
  title={FinDeepForecast: A Live Multi-Agent System for Benchmarking Deep Research Agents in Financial Forecasting},
  author={Li, Xiangyu and Yao, Xuan and Qi, Guohao and Zhu, Fengbin and Koa, Kelvin JL and Ng, Xiang Yao and Liu, Ziyang and Ni, Xingyu and Liu, Chang and Yang, Yonghui and others},
  journal={arXiv preprint arXiv:2601.05039},
  year={2026}
}

@inproceedings{siru2025time,
  title={Time-VLM: Exploring Multimodal Vision-Language Models for Augmented Time Series Forecasting},
  author={Siru, Zhong and Weilin, Ruan and Jin, Ming and Huan, Li and Qingsong, Wen and Yuxuan, Liang},
  booktitle={Forty-Second International Conference on Machine Learning (ICML 2025)},
  year={2025},
  organization={Proceedings of Machine Learning Research}
}

@article{zhao2025timeseriesscientist,
  title={Timeseriesscientist: A general-purpose ai agent for time series analysis},
  author={Zhao, Haokun and Zhang, Xiang and Wei, Jiaqi and Xu, Yiwei and He, Yuting and Sun, Siqi and You, Chenyu},
  journal={arXiv preprint arXiv:2510.01538},
  year={2025}
}

@inproceedings{ning2025tsrag,
  title={TS-RAG: Retrieval-Augmented Generation based Time Series Foundation Models are Stronger Zero-Shot Forecaster},
  author={Ning, Kanghui and Pan, Zijie and Liu, Yu and Jiang, Yushan and Zhang, James Y. and Rasul, Kashif and Schneider, Anderson and Ma, Lintao and Nevmyvaka, Yuriy and Song, Dongjin},
  booktitle={Advances in Neural Information Processing Systems (NeurIPS)},
  year={2025}
}

@inproceedings{strong2025tsver,
  title={TSVer: A Benchmark for Fact Verification Against Time-Series Evidence},
  author={Strong, Marek and Vlachos, Andreas},
  booktitle={Proceedings of the 2025 Conference on Empirical Methods in Natural Language Processing},
  pages={29894--29914},
  year={2025}
}

@article{yang2025time,
  title={Time-ra: Towards time series reasoning for anomaly with llm feedback},
  author={Yang, Yiyuan and Liu, Zichuan and Song, Lei and Ying, Kai and Wang, Zhiguang and Bamford, Tom and Vyetrenko, Svitlana and Bian, Jiang and Wen, Qingsong},
  journal={arXiv preprint arXiv:2507.15066},
  year={2025}
}

@article{parker2025eliciting,
  title={Eliciting chain-of-thought reasoning for time series analysis using reinforcement learning},
  author={Parker, Felix and Chan, Nimeesha and Zhang, Chi and Ghobadi, Kimia},
  journal={arXiv preprint arXiv:2510.01116},
  year={2025}
}

@article{yu2025ts,
  title={TS-Reasoner: Aligning Time Series Foundation Models with LLM Reasoning},
  author={Yu, Fangxu and Zhao, Hongyu and Zhou, Tianyi},
  journal={arXiv preprint arXiv:2510.03519},
  year={2025}
}

@article{cao2025conversational,
  title={Conversational Time Series Foundation Models: Towards Explainable and Effective Forecasting},
  author={Cao, Defu and Gee, Michael and Liu, Jinbo and Wang, Hengxuan and Yang, Wei and Wang, Rui and Liu, Yan},
  journal={arXiv preprint arXiv:2512.16022},
  year={2025}
}

@inbook{10.1145/3768292.3771251,
author = {Ang, Yihao and Bao, Yifan and Jiang, Lei and Tao, Jiajie and Tung, Anthony K. H. and Szpurch, Lukasz and Ni, Hao},
title = {Structured Agentic Workflows for Financial Time-Series Modelling with LLMs and Reflective Feedback},
year = {2025},
isbn = {9798400722202},
publisher = {Association for Computing Machinery},
address = {New York, NY, USA},
booktitle = {Proceedings of the 6th ACM International Conference on AI in Finance},
pages = {924–932},
numpages = {9}
}

@article{yeh2025empowering,
  title={Empowering Time Series Forecasting with LLM-Agents},
  author={Yeh, Chin-Chia Michael and Lai, Vivian and Saini, Uday Singh and Fan, Xiran and Fan, Yujie and Wang, Junpeng and Dai, Xin and Zheng, Yan},
  journal={arXiv preprint arXiv:2508.04231},
  year={2025}
}

@article{bazaga2025learning,
  title={Learning to Reason Over Time: Timeline Self-Reflection for Improved Temporal Reasoning in Language Models},
  author={Bazaga, Adri{\'a}n and Blloshmi, Rexhina and Byrne, Bill and de Gispert, Adri{\`a}},
  journal={arXiv preprint arXiv:2504.05258},
  year={2025}
}

@article{su2025chain,
  title={Chain-of-thought Reviewing and Correction for Time Series Question Answering},
  author={Su, Chen and Tian, Yuanhe and Song, Yan},
  journal={arXiv preprint arXiv:2512.22627},
  year={2025}
}

@article{zhang2025can,
  title={Can Competition Enhance the Proficiency of Agents Powered by Large Language Models in the Realm of News-driven Time Series Forecasting?},
  author={Zhang, Yuxuan and Feng, Yangyang and Li, Daifeng and Zhang, Kexin and Chen, Junlan and Deng, Bowen},
  journal={arXiv preprint arXiv:2504.10210},
  year={2025}
}

@article{zhou2025enhancing,
  title={Enhancing LLM Reasoning for Time Series Classification by Tailored Thinking and Fused Decision},
  author={Zhou, Jiahui and Li, Dan and Li, Lin and Chen, Zhuomin and Wu, Shunyu and Ye, Haozheng and Lou, Jian and Spanos, Costas J},
  journal={arXiv preprint arXiv:2506.00807},
  year={2025}
}

@inproceedings{jalori-etal-2025-flairr,
    title = "{FLAIRR}-{TS} - Forecasting {LLM}-Agents with Iterative Refinement and Retrieval for Time Series",
    author = "Jalori, Gunjan  and
      Verma, Preetika  and
      Arik, Sercan O",
    editor = "Christodoulopoulos, Christos  and
      Chakraborty, Tanmoy  and
      Rose, Carolyn  and
      Peng, Violet",
    booktitle = "Findings of the Association for Computational Linguistics: EMNLP 2025",
    month = nov,
    year = "2025",
    address = "Suzhou, China",
    publisher = "Association for Computational Linguistics",
    url = "https://aclanthology.org/2025.findings-emnlp.834/",
    doi = "10.18653/v1/2025.findings-emnlp.834",
    pages = "15427--15437",
    ISBN = "979-8-89176-335-7"
}

@article{liu2026rationale,
  title={Rationale-Grounded In-Context Learning for Time Series Reasoning with Multimodal Large Language Models},
  author={Liu, Qingxiang and Cui, Zhiqing and Luo, Xiaoliang and Wu, Yuqian and Jiang, Zhuoyang and Wan, Huaiyu and Sun, Sheng and Wang, Lvchun and Yu, Wei and Liang, Yuxuan},
  journal={arXiv preprint arXiv:2601.02968},
  year={2026}
}

@inproceedings{
wang2025itformer,
title={{ITF}ormer: Bridging Time Series and Natural Language for Multi-Modal {QA} with Large-Scale Multitask Dataset},
author={Yilin wang and Peixuan Lei and Jie Song and Yuzhe Hao and Tao Chen and Yuxuan Zhang and LEI JIA and Yuanxiang Li and zhongyu wei},
booktitle={Forty-second International Conference on Machine Learning},
year={2025},
url={https://openreview.net/forum?id=GByP03IitA}
}

@article{he2025radarqa,
  title={Radarqa: Multi-modal quality analysis of weather radar forecasts},
  author={He, Xuming and You, Zhiyuan and Gong, Junchao and Liu, Couhua and Yue, Xiaoyu and Zhuang, Peiqin and Zhang, Wenlong and Bai, Lei},
  journal={arXiv preprint arXiv:2508.12291},
  year={2025}
}

@article{park2026bridging,
  title={Bridging Temporal and Textual Modalities: A Multimodal Framework for Automated Cloud Failure Root Cause Analysis},
  author={Park, Gijun},
  journal={arXiv preprint arXiv:2601.04709},
  year={2026}
}

@article{cui2025augur,
  title={Augur: Modeling Covariate Causal Associations in Time Series via Large Language Models},
  author={Cui, Zhiqing and Wang, Binwu and Liu, Qingxiang and Wang, Yeqiang and Zhou, Zhengyang and Liang, Yuxuan and Wang, Yang},
  journal={arXiv preprint arXiv:2510.07858},
  year={2025}
}

@article{sun2025timemkg,
  title={TimeMKG: Knowledge-Infused Causal Reasoning for Multivariate Time Series Modeling},
  author={Sun, Yifei and Liu, Junming and Chen, Yirong and Yan, Xuefeng and Wang, Ding},
  journal={arXiv preprint arXiv:2508.09630},
  year={2025}
}

@article{chukwu2025counterfactual,
  title={Counterfactual Explanations for Time Series Should be Human-Centered and Temporally Coherent in Interventions},
  author={Chukwu, Emmanuel C and Schouten, Rianne M and Tabak, Monique and Pechenizkiy, Mykola},
  journal={arXiv preprint arXiv:2512.14559},
  year={2025}
}

@article{rodling2025causal,
  title={Causal Explanations Over Time: Articulated Reasoning for Interactive Environments},
  author={R{\"o}dling, Sebastian and Ze{\v{c}}evi{\'c}, Matej and Dhami, Devendra Singh and Kersting, Kristian},
  journal={arXiv preprint arXiv:2506.03915},
  year={2025}
}

@inproceedings{zhang2025camef,
  title={CAMEF: Causal-augmented multi-modality event-driven financial forecasting by integrating time series patterns and salient macroeconomic announcements},
  author={Zhang, Yang and Yang, Wenbo and Wang, Jun and Ma, Qiang and Xiong, Jie},
  booktitle={Proceedings of the 31st ACM SIGKDD Conference on Knowledge Discovery and Data Mining V. 2},
  pages={3867--3878},
  year={2025}
}

@article{cerutti2025methodological,
  title={Methodological Insights into Structural Causal Modelling and Uncertainty-Aware Forecasting for Economic Indicators},
  author={Cerutti, Federico},
  journal={arXiv preprint arXiv:2509.07036},
  year={2025}
}

@article{niu2025event,
  title={Event-CausNet: Unlocking Causal Knowledge from Text with Large Language Models for Reliable Spatio-Temporal Forecasting},
  author={Niu, Luyao and Wang, Zepu and Guan, Shuyi and Liu, Yang and Sun, Peng},
  journal={arXiv preprint arXiv:2511.12769},
  year={2025}
}

@article{zhang2025medkgent,
  title={MedKGent: A Large Language Model Agent Framework for Constructing Temporally Evolving Medical Knowledge Graph},
  author={Zhang, Duzhen and Wang, Zixiao and Li, Zhong-Zhi and Yu, Yahan and Jia, Shuncheng and Dong, Jiahua and Xu, Haotian and Wu, Xing and Zhang, Yingying and Zhang, Tielin and others},
  journal={arXiv preprint arXiv:2508.12393},
  year={2025}
}

@article{wu2025scits,
  title={SciTS: Scientific Time Series Understanding and Generation with LLMs},
  author={Wu, Wen and Zhang, Ziyang and Liu, Liwei and Xu, Xuenan and Liu, Junlin and Fan, Ke and Lv, Qitan and Zhuang, Jimin and Zhang, Chen and Yuan, Zheqi and others},
  journal={arXiv preprint arXiv:2510.03255},
  year={2025}
}

@InProceedings{pmlr-v267-li25ah,
  title = 	 {{BRIDGE}: Bootstrapping Text to Control Time-Series Generation via Multi-Agent Iterative Optimization and Diffusion Modeling},
  author =       {Li, Hao and Huang, Yu-Hao and Xu, Chang and Schlegel, Viktor and Jiang, Renhe and Batista-Navarro, Riza and Nenadic, Goran and Bian, Jiang},
  booktitle = 	 {Proceedings of the 42nd International Conference on Machine Learning},
  pages = 	 {34742--34773},
  year = 	 {2025},
  editor = 	 {Singh, Aarti and Fazel, Maryam and Hsu, Daniel and Lacoste-Julien, Simon and Berkenkamp, Felix and Maharaj, Tegan and Wagstaff, Kiri and Zhu, Jerry},
  volume = 	 {267},
  series = 	 {Proceedings of Machine Learning Research},
  month = 	 {13--19 Jul},
  publisher =    {PMLR},
  url = 	 {https://proceedings.mlr.press/v267/li25ah.html}
}

@article{fuest2025cents,
  title={CENTS: Generating synthetic electricity consumption time series for rare and unseen scenarios},
  author={Fuest, Michael and Cuesta, Alfredo and Veeramachaneni, Kalyan},
  journal={arXiv preprint arXiv:2501.14426},
  year={2025}
}

@misc{li2025timetravel,
      title={Time Travel is Cheating: Going Live with DeepFund for Real-Time Fund Investment Benchmarking}, 
      author={Changlun Li and Yao Shi and Chen Wang and Qiqi Duan and Runke Ruan and Weijie Huang and Haonan Long and Lijun Huang and Nan Tang and Yuyu Luo},
      year={2025},
      eprint={2505.11065},
      archivePrefix={arXiv},
      primaryClass={cs.CE},
      url={https://arxiv.org/abs/2505.11065}, 
}

@article{lan2025gem,
  title={Gem: Empowering mllm for grounded ecg understanding with time series and images},
  author={Lan, Xiang and Wu, Feng and He, Kai and Zhao, Qinghao and Hong, Shenda and Feng, Mengling},
  journal={arXiv preprint arXiv:2503.06073},
  year={2025}
}

@article{li2025zara,
  title={Zara: Zero-shot motion time-series analysis via knowledge and retrieval driven llm agents},
  author={Li, Zechen and Chen, Baiyu and Xue, Hao and Salim, Flora D},
  journal={arXiv preprint arXiv:2508.04038},
  year={2025}
}

@article{zhang2025timemaster,
  title={TimeMaster: Training Time-Series Multimodal LLMs to Reason via Reinforcement Learning},
  author={Zhang, Junru and Feng, Lang and Guo, Xu and Wu, Yuhan and Dong, Yabo and Xu, Duanqing},
  journal={arXiv preprint arXiv:2506.13705},
  year={2025}
}

@article{wang2026slep,
  title   = {Large-model-based smart agent for time series anomaly detection in power systems},
  author  = {Wang, Bingrui and Zhou, Yuan and Ge, Leijiao and Kung, Sun-Yuan},
  journal = {Expert Systems with Applications},
  volume  = {296},
  number  = {Part B},
  pages   = {128917},
  year    = {2026},
  issn    = {0957-4174},
  doi     = {10.1016/j.eswa.2025.128917},
  url     = {https://www.sciencedirect.com/science/article/pii/S0957417425025345}
}

@inproceedings{kong2025fusing,
  title={Fusing Narrative Semantics for Financial Volatility Forecasting},
  author={Kong, Yaxuan and Hwang, Yoontae and Kaiser, Marcus and Vryonides, Chris and Oomen, Roel and Zohren, Stefan},
  booktitle={Proceedings of the 6th ACM International Conference on AI in Finance},
  pages={683--691},
  year={2025}
}

@article{hwang2025decision,
  title={Decision-informed neural networks with large language model integration for portfolio optimization},
  author={Hwang, Yoontae and Kong, Yaxuan and Zohren, Stefan and Lee, Yongjae},
  journal={arXiv preprint arXiv:2502.00828},
  year={2025}
}

@inproceedings{
liu2025timerxl,
title={Timer-{XL}: Long-Context Transformers for Unified Time Series Forecasting},
author={Yong Liu and Guo Qin and Xiangdong Huang and Jianmin Wang and Mingsheng Long},
booktitle={The Thirteenth International Conference on Learning Representations},
year={2025},
url={https://openreview.net/forum?id=KMCJXjlDDr}
}

@article{cai2025timeseriesgym,
  title={TimeSeriesGym: A Scalable Benchmark for (Time Series) Machine Learning Engineering Agents},
  author={Cai, Yifu and Li, Xinyu and Goswami, Mononito and Wili{\'n}ski, Micha{\l} and Welter, Gus and Dubrawski, Artur},
  journal={arXiv preprint arXiv:2505.13291},
  year={2025}
}

@article{chang2025time,
  title={Time-IMM: A Dataset and Benchmark for Irregular Multimodal Multivariate Time Series},
  author={Chang, Ching and Hwang, Jeehyun and Shi, Yidan and Wang, Haixin and Peng, Wen-Chih and Chen, Tien-Fu and Wang, Wei},
  journal={arXiv preprint arXiv:2506.10412},
  year={2025}
}

@article{liu2025evaluating,
  title={Evaluating system 1 vs. 2 reasoning approaches for zero-shot time series forecasting: A benchmark and insights},
  author={Liu, Haoxin and Zhao, Zhiyuan and Li, Shiduo and Prakash, B Aditya},
  journal={arXiv preprint arXiv:2503.01895},
  year={2025}
}

@article{reuters_netflix_wbd_deal_2025,
  author  = {{Reuters}},
  title   = {Netflix to buy Warner Bros Discovery's studios, streaming unit for \$72 billion},
  journal = {Reuters},
  year    = {2025},
  month   = dec,
  day     = {5},
  url     = {https://www.reuters.com/legal/transactional/netflix-agrees-buy-warner-bros-discoverys-studios-streaming-division-2025-12-05/},
  note    = {Deal described as about \$72B equity value and \$82.7B including debt (often rounded to \textasciitilde\$83B)}
}

@article{reuters_netflix_earnings_2026,
  author  = {{Reuters}},
  title   = {Netflix slightly beats revenue estimates, shares slide amid bidding war for Warner Bros},
  journal = {Reuters},
  year    = {2026},
  month   = jan,
  day     = {20},
  url     = {https://www.reuters.com/business/media-telecom/netflix-beats-revenue-estimates-subscribers-reach-325-million-2026-01-20/},
  note    = {Reports Q4 revenue about \$12.1B, EPS 56 cents vs. 55-cent estimate, and 325M paid subscribers; notes shares down in after-hours}
}

@article{reuters_netflix_all_cash_2026,
  author  = {Chmielewski, Dawn},
  title   = {Netflix will now pay all cash for Warner Bros to keep Paramount at bay},
  journal = {Reuters},
  year    = {2026},
  month   = jan,
  day     = {21},
  url     = {https://www.reuters.com/business/finance/netflix-submits-amended-all-cash-offer-warner-bros-wins-board-support-2026-01-20/},
  note    = {States Netflix shares have dropped about/almost 15\% since the Dec. 5 merger announcement}
}

@article{fortune_netflix_earnings_2026,
  author  = {{Fortune}},
  title   = {Netflix stock sinks after earnings call, as co-CEOs can't convince investors about Warner Bros deal},
  journal = {Fortune},
  year    = {2026},
  month   = jan,
  day     = {20},
  url     = {https://fortune.com/2026/01/20/netflix-earnings-stock-falls-warner-brothers-bid-stock-buybacks/},
  note    = {Reports stock down 4.9\% after-hours; discusses investor concern and buyback pause to help finance the deal}
}

@misc{netflix_shareholder_letter_q4_2025,
  author       = {{Netflix, Inc.}},
  title        = {Q4 2025 Shareholder Letter},
  year         = {2026},
  month        = jan,
  day          = {20},
  howpublished = {\url{https://s22.q4cdn.com/959853165/files/doc_financials/2025/q4/FINAL-Q4-25-Shareholder-Letter.pdf}},
  note = {Reports revenue \$12.051B (+17.6\% YoY, often rounded to 18\%), diluted EPS \$0.56, and 325M+ paid memberships; includes disclosure on share repurchases/buybacks}
}

@article{apnews_wbd_paramount_vs_netflix_2025,
  author  = {{Associated Press}},
  title   = {Warner Bros. urges shareholders to reject Paramount bid in favor of Netflix's},
  journal = {Associated Press},
  year    = {2025},
  month   = dec,
  url     = {https://apnews.com/article/
             d025a585f7a77cb9d8b066e65576101f},
  note    = {Covers Paramount Skydance rival bid context and Warner board recommendation dynamics}
}
\bibliographystyle{icml2026}

\newpage
\appendix
\onecolumn

\section{Motivation - Case Study}
\label{appendix_case_study}
\paragraph{Background.}
In early December 2025, Netflix announced a proposed acquisition of Warner Bros.\ Discovery valued at approximately \$83 billion, representing one of the largest media acquisitions in recent history \cite{reuters_netflix_wbd_deal_2025}. However, Netflix's stock price experienced significant declines following the announcement, dropping approximately 15\% from early December through January 20, 2026 \cite{reuters_netflix_all_cash_2026}. On January 20, 2026, Netflix released its Q4 2025 earnings report, reporting revenue of \$12.05 billion (up 18\% year-over-year), diluted EPS of \$0.56 (beating estimates by \$0.01), and crossing 325 million paid subscribers \cite{netflix_shareholder_letter_q4_2025,reuters_netflix_earnings_2026}. Despite these positive results, following the earnings call where co-CEOs Ted Sarandos and Greg Peters attempted to reassure investors about the Warner Bros.\ acquisition, the stock fell an additional 4.9\% in after-hours trading \cite{fortune_netflix_earnings_2026}. The stock price decline occurred despite several positive financial indicators, with investors expressing concerns about Netflix's decision to temporarily halt share buybacks to finance the acquisition, fears about management distraction, and competition from a rival bid by Paramount Skydance \cite{netflix_shareholder_letter_q4_2025,fortune_netflix_earnings_2026,apnews_wbd_paramount_vs_netflix_2025}.

\paragraph{The Task.}
This scenario presents a complex time series reasoning task evaluated through specific questions that test comprehensive analysis across multiple dimensions. The following questions illustrate the requirements:

\begin{tcolorbox}[colback=gray!5!white,colframe=gray!75!black,title=Question 1: Comprehensive Stock Price Feature Analysis]
\textbf{Task:} Analyze Netflix's stock price data from December 1, 2025 through January 20, 2026, including open, high, low, close, and adjusted close prices. Calculate and analyze realized volatility (daily, weekly, rolling windows), intraday volatility patterns, and identify volatility regime changes. Examine trading volume patterns, volume-weighted average price (VWAP), and volume trends relative to price movements. \\
\textbf{Question:} What temporal characteristics distinguish this period from Netflix's historical performance across these dimensions?
\end{tcolorbox}

\begin{tcolorbox}[colback=gray!5!white,colframe=gray!75!black,title=Question 2: Earnings Report Integration]
\textbf{Task:} Retrieve and analyze Netflix's Q4 2025 earnings report released on January 20, 2026, including the earnings call transcript, financial statements, and key metrics (revenue \$12.05B, EPS \$0.56, 325M subscribers). \\
\textbf{Questions:} How do these earnings results compare to analyst expectations and prior quarters? What specific guidance or forward-looking statements were made during the earnings call? How do these earnings-related factors temporally align with stock price movements?
\end{tcolorbox}

\begin{tcolorbox}[colback=gray!5!white,colframe=gray!75!black,title=Question 3: Multi-Source Context Integration]
\textbf{Task:} Retrieve sell-side analyst reports from major investment banks covering Netflix during December 2025 through January 2026. Identify price target revisions, rating changes (upgrades/downgrades), and key analyst concerns or recommendations. Retrieve and synthesize relevant news articles from financial news outlets (Bloomberg, Reuters, Financial Times, Wall Street Journal) and industry publications. Analyze Twitter/X discussions, Reddit sentiment (r/stocks, r/investing), and social media volume related to Netflix. \\
\textbf{Question:} How did analyst sentiment, news coverage, and social media sentiment evolve following the Warner Bros. acquisition announcement and the Q4 earnings release?
\end{tcolorbox}

\begin{tcolorbox}[colback=gray!5!white,colframe=gray!75!black,title=Question 4: Causal Reasoning with Evidence]
\textbf{Task:} Explain the causal relationship between Netflix's Warner Bros. acquisition announcement and its stock price decline, incorporating evidence from earnings reports, analyst reports, news coverage, and social media sentiment. Distinguish between firm-specific factors (acquisition concerns, competitive bidding, earnings-related factors) and broader market conditions. \\
\textbf{Questions:} What specific evidence from each information source supports or contradicts each potential explanation? How do volatility and volume patterns support your causal analysis?
\end{tcolorbox}

\begin{tcolorbox}[colback=gray!5!white,colframe=gray!75!black,title=Question 5: Evidence Verification and Time Alignment]
\textbf{Task:} Verify your explanation by cross-referencing specific dates and times. \\
\textbf{Questions:} When was the acquisition announced (exact date and time)? When did Netflix report Q4 2025 earnings (exact date and time)? When did key analyst reports get published? When did major news articles appear? When did social media discussions peak? When did specific stock price declines, volatility spikes, and volume surges occur? Are all these events correctly time-aligned in your analysis? Identify any temporal misalignments or gaps in your evidence chain.
\end{tcolorbox}

\begin{tcolorbox}[colback=gray!5!white,colframe=gray!75!black,title=Question 6: Decision-Making Recommendation]
\textbf{Task:} Based on your comprehensive analysis incorporating stock price features, volatility, volume, earnings reports, analyst reports, news articles, and social media sentiment. \\
\textbf{Question:} Should an investor buy, hold, or sell Netflix stock as of January 20, 2026? Provide a recommendation with detailed reasoning that synthesizes all information sources, temporal patterns, external context, and causal relationships. What are the key risk factors and opportunities identified across all information sources?
\end{tcolorbox}

This task requires open-ended reasoning, comprehensive multimodal integration, and evidence-grounded decision-making that goes far beyond traditional forecasting or classification tasks.

\paragraph{Insights and Gaps.}
When examined using this Netflix case study, we had a detailed look at Time-MQA~\cite{kong2025time} and GPT-5.2-Auto, which helped us identify several critical gaps illustrated in our analysis. 

\textbf{Gap \#1 (Model Design):} Time-MQA cannot retrieve external context (earnings reports, analyst reports, news articles, social media) and lacks web search capabilities. GPT-5.2-Auto demonstrates practical challenges: it is slow, automatically enters extended \enquote{thinking mode} that delays responses, defaults to generating code rather than performing direct time series reasoning, and could only partially satisfy the request (retrieving data only for December 29, 2025 through January 20, 2026, missing the first three weeks of December). The model's response was heavily code-focused, computing technical indicators through programming rather than reasoning about temporal patterns and their relationships to external events. Neither model explicitly models causal reasoning structures that connect stock price features, volatility, and volume with earnings reports, analyst reports, and social media sentiment, nor do they provide transparent reasoning paths or verification mechanisms. 

\textbf{Gap \#2 (Task Formulation):} The Netflix scenario requires causal analysis, counterfactual reasoning, and decision-making tasks that go beyond traditional forecasting or classification, demanding synthesis of multiple information sources without predefined answer choices, capabilities that current templated QA formats cannot support. 

\textbf{Gap \#3 (Evaluation):} This real-world scenario tests models on actual market events with complex, ambiguous causal relationships across multiple information sources, requiring multi-step reasoning, evidence grounding with specific dates and times, and handling conflicting information, capabilities rarely evaluated in current benchmarks. This study demonstrates that achieving reliable time series reasoning for real-world decision-making requires addressing these three gaps.

\section{Background}
\label{appendix_background}

\subsection{Multimodal Large Language Models (MLLMs)}

A Multimodal Large Language Model (MLLM) is an advanced AI system that extends the reasoning capabilities of Large Language Models (LLMs) by enabling them to process, interpret, and generate information across multiple modalities, including text, images, audio, and time series data~\cite{wu2023multimodal}. Unlike traditional LLMs that rely solely on textual data, MLLMs integrate multimodal representations through sophisticated deep-learning architectures, allowing them to perceive and reason about complex relationships between different data types. This capability enhances their performance in tasks such as multimodal question answering, image and video captioning, and medical image analysis, where understanding information from multiple sources is essential~\cite{yin2023survey}. By leveraging advanced fusion mechanisms, MLLMs generate contextually rich and coherent outputs that go beyond text-based reasoning, making them highly effective in applications requiring comprehensive multimodal understanding~\cite{zhang2024mm}.

\subsection{Multimodal Time Series}
The increasing availability of heterogeneous data has highlighted the need to better capture the complexity of real-world phenomena. Multimodal time series address this challenge by integrating data from multiple modalities, where each modality represents a distinct type of information, such as images, text, audio, or structured numerical data. By extending traditional single-modal time series analysis, this approach enables the incorporation of diverse and complementary data sources, providing a more comprehensive understanding of complex systems. Formally, a multimodal time series can be represented as $\mathbf{X} = {\mathbf{X}^{(m)}}_{m=1}^M$, where $M$ denotes the total number of modalities, and each $\mathbf{X}^{(m)}$ corresponds to the time series for modality $m$. For instance, $\mathbf{X}^{(m)} = (\mathbf{x}_0^{(m)}, \mathbf{x}_1^{(m)}, \dots, \mathbf{x}_{T-1}^{(m)})$ represents the sequential data of modality $m$ over $T$ time steps, with $\mathbf{x}_t^{(m)}$ varying in form depending on the modality, such as vectors for numerical data, matrices for images, or sequences for text and audio. Multimodal time series analysis enables a more comprehensive understanding of temporal patterns and interactions across modalities by integrating multiple data sources. This is particularly important in applications such as healthcare, autonomous driving, and multimedia analysis~\cite{moor2023foundation,cui2024survey,zhou2024survey}.

\subsection{Time Series Classical Tasks}
Time series analysis and its various classical tasks are widely applied across real-world domains, such as financial forecasting, healthcare monitoring, traffic flow analysis, climate modeling, industrial predictive maintenance, and AIOps~\cite{shumway2000time,nie2024survey,yang2021early,yang2023sgdp,liang2024foundation}. Also, time series analysis encompasses a diverse set of tasks aimed at extracting insights and addressing challenges in temporal data~\cite{hamilton2020time,kirchgassner2012introduction}. Among them, common tasks include forecasting, which predicts future values based on historical trends and can be divided into short-term and long-term predictions~\cite{wen2022transformers}, and anomaly detection, which identifies unusual patterns or deviations from expected behavior~\cite{zamanzadeh2024deep}. Imputation addresses missing or corrupted data points to ensure dataset completeness~\cite{du2024tsi}, while generation creates synthetic time series to replicate statistical properties for data augmentation or scenario simulation~\cite{yang2024survey}. Other tasks include classification, which assigns categorical labels based on patterns, and regression, which predicts continuous target values~\cite{mohammadi2024deep,yang2021long}. In recent years, an increasing number of approaches have explored leveraging multimodal data to enhance classical time series analysis tasks, validating the effectiveness and rationality of these methods~\cite{zhou2023one,liutime,gruver2024large,chang2023llm4ts,cao2023tempo,yin2023survey}. For instance, Time-LLM aligns and reprograms LLMs for time series forecasting through textual input alignment~\cite{jin2024time}, while Time-MMD incorporates additional textual data with Transformer-based models using weighted fusion to perform time series forecasting and potentially other tasks~\cite{liutime}. Beyond text modalities, medical data employs supplementary image or tabular data to enhance time series analysis using foundation models~\cite{hollmann2025accurate,moor2023foundation}. Moreover, methods leveraging generative models, such as diffusion models, have been proposed. These approaches inject multiple modalities into the conditional space, improving the robustness of temporal tasks and enhancing generative diversity~\cite{yang2024survey}.

\section{Related Works} 
We conducted a comprehensive review and investigation of trends in Time Series Reasoning research spanning the period from 2023 to 2025. Our analysis reveals a rapid growth trajectory: 7 papers were published in 2023, 69 papers in 2024, and 580 papers in 2025, totaling 656 papers across the three-year period. This exponential growth reflects the emerging and rapidly evolving nature of time series reasoning as a research area.






To facilitate systematic analysis, we employed a rule-based categorization approach using keyword matching to classify papers into distinct research categories. Our categorization framework identified the following research areas: General / Methodology (325 papers, 49.5\%), General - Video (60 papers, 9.1\%), General - Agents (41 papers, 6.2\%), General - Time Series Reasoning (26 papers, 4.0\%), General - Reasoning (23 papers, 3.5\%), General - Temporal Reasoning (18 papers, 2.7\%), General - Graph (10 papers, 1.5\%), Finance \& Economy (44 papers, 6.7\%), Healthcare \& Bio (37 papers, 5.6\%), Traffic \& Mobility (21 papers, 3.2\%), Energy \& Power (20 papers, 3.0\%), Industrial \& IoT (18 papers, 2.7\%), IT \& Operations (7 papers, 1.1\%), and Climate \& Environment (6 papers, 0.9\%). All categorizations were human-verified to ensure accuracy and consistency. This distribution highlights the emerging nature of time series reasoning research, with a significant portion of work appearing as the field rapidly develops.

To support ongoing research and enable comprehensive exploration of the collected papers, we have developed an interactive dashboard that provides researchers with flexible filtering, search, and analysis capabilities across multiple dimensions such as research category, application domain, and publication year. This tool facilitates the discovery of relevant work and identification of research trends. It is available at \url{https://github.com/Eleanorkong/Awesome-Time-Series-Reasoning}.

Below, we summarize the latest work in two key areas: (1) Reasoning in NLP, and (2) Reasoning in Time Series.

\label{appendix_related_works}
\subsection{Reasoning in NLP}
In the field of Natural Language Processing (NLP), reasoning refers to the process of deriving conclusions from textual evidence and logical principles~\cite{besta2025reasoning}. It involves tasks such as understanding implicit information, performing logical inferences, and applying commonsense knowledge. Reasoning capabilities are crucial for addressing complex language tasks like natural language inference, multi-hop question answering, and commonsense reasoning~\cite{yu2024natural}. Types of reasoning include Chain-of-Thought (CoT), which breaks problems into intermediate steps for clarity, deductive reasoning which applies general rules to specific cases, and inductive reasoning which generalizes from observations~\cite{xia2024beyond}. Abductive reasoning identifies the most plausible explanations, while analogical reasoning transfers knowledge based on similarities~\cite{shi2024language,lewis2024using}. Others include commonsense reasoning, probabilistic reasoning, and causal reasoning~\cite{yu2024natural}. These approaches enhance interpretability of NLP systems. 

\subsection{Reasoning in Time Series}
Time series analysis tasks traditionally focus on narrower objectives — like forecasting or anomaly detection — each addressed by its own specialized model, often relying solely on numerical patterns within the data. In contrast, time series reasoning with logic integrates multiple tasks under a single, context-aware framework with human-like reasoning~\cite{chow2024towards}. It readily incorporates domain knowledge and external data sources, providing natural language explanations and causal insights rather than mere numerical outputs~\cite{potosnak2024implicit}. This approach allows time series reasoning to adapt to shifting conditions and novel questions, delving into the “why” behind observed patterns and bridging the gap between automated analysis and real-world decision-making. Furthermore, reasoning in NLP can enhance time series analysis by enabling models to infer complex temporal patterns and relationships, improving interpretability and decision-making~\cite{potosnak2024implicit,cai2024timeseriesexam}. By incorporating reasoning capabilities, models can better handle ambiguous or incomplete data, leading to more robust predictions and insights. 

Our review of work from 2023 to 2025 indicates that the field has grown rapidly: we identified 656 papers at the intersection of time series and reasoning/LLMs, including 67 papers explicitly focused on time series reasoning, temporal reasoning, or general reasoning applied to time series. Research spans foundation models and methodology, domain-specific applications (e.g., finance, healthcare, traffic, energy), agent-based systems, and evaluation benchmarks. To gain deeper insights into the current state of the field, we conducted a detailed analysis of 20 influential papers from 2025, examining them across three critical dimensions: model design (representation, knowledge integration, reasoning mechanisms, and validation), task formulation (scope and complexity), and evaluation (reasoning quality and real-world applicability). These papers include foundational works on multimodal time series understanding (ChatTime~\cite{wang2025chattime}, ChatTS~\cite{xie2025chatts}, ITFormer~\cite{wang2025itformer}), question answering frameworks (Time-MQA~\cite{kong2025time}), agent-based systems (TimeSeriesScientist~\cite{zhao2025timeseriesscientist}, TimeCAP~\cite{lee2025timecap}), reasoning-enhanced models (TimeOmni-1~\cite{guan2025timeomni}, TimeMaster~\cite{zhang2025timemaster}, MERIT~\cite{zhou2025enhancing}), anomaly detection (Zhou and Yu~\cite{zhou2024can}), visualization-based reasoning (Liu et al.~\cite{liu2025picture}), explainable systems (TimeXL~\cite{liu2025timerxl}), benchmarks and evaluation (Context is Key~\cite{williams2024context}, TimeSeriesGym~\cite{cai2025timeseriesgym}, Time-IMM~\cite{chang2025time}, Evaluating System 1 vs. 2~\cite{liu2025evaluating}), domain-specific applications (OpenTSLM~\cite{langer2025opentslm}, GEM~\cite{lan2025gem}), and real-world deployment (Time Travel is Cheating~\cite{li2025timetravel}). This comprehensive analysis revealed significant gaps in model design (limited dynamic contextual guidance and verification mechanisms), task formulation (narrow scope compared to real-world needs), and evaluation (insufficient assessment of reasoning quality and decision-making readiness), as detailed in our paper.

Recent advances in 2025 have particularly emphasized multimodal integration, with several works bridging numerical time series data with textual information, images, and external knowledge sources. Agent-based frameworks have emerged as a promising direction, enabling autonomous analysis workflows that can retrieve relevant context, reason through complex scenarios, and provide explainable outputs. Reinforcement learning has been increasingly adopted to enhance reasoning capabilities. Despite this rapid growth and progress, open challenges remain including optimal representation of temporal data for LLMs, effective training paradigms, dynamic contextual guidance with time-aligned verification, and comprehensive evaluation of reasoning quality and decision-making readiness. Reasoning in time series analysis thus remains a highly active and impactful area with significant room for further development~\cite{chow2024towards, cao2025conversational}.

\section{Gap \#1: Model Design} 
\label{appendix_model_design}

\subsection{Reasoning Structure} 
\label{appendix_reasoning_structure}
We summarize reasoning structure types in Table~\ref{tab:reasoning_types}. It includes four types: End-to-end Reasoning, Forward Reasoning, Backward Reasoning, and Forward-Backward Reasoning. End-to-end reasoning is characterized by its direct mapping from inputs to outputs, encapsulating the reasoning process within hidden states, which makes it less interpretable but effective for tasks requiring concise outputs. Forward Reasoning, on the other hand, adopts a bottom-up approach, explicitly stating intermediate steps, making it suitable for tasks like solving math problems or predicting time series trends sequentially. Backward Reasoning employs a top-down strategy, breaking down the main problem into smaller sub-problems, which is particularly useful for diagnostic tasks such as identifying the causes of anomalies. Lastly, Forward-Backward Reasoning combines forward and backward approaches, proposing potential solutions and verifying them, making it ideal for complex tasks like analyzing time series anomalies. 

Moreover, these structures can be further organized into chain-, tree-, or graph-based formalisms to represent the reasoning path more explicitly. A \textit{chain-based} approach arranges the reasoning steps in a sequential, linear fashion, making it straightforward to follow how each conclusion is derived. A \textit{tree-based} approach expands reasoning into branches, allowing multiple concurrent paths and a hierarchical breakdown of complex problems. Meanwhile, a \textit{graph-based} approach generalizes these connections and can account for interdependencies and cyclic references among different inference steps. 

\begin{table*}[t]
    \caption{Comparison of reasoning structure types with definitions and examples.}
    \vspace{-1mm}
    \centering
    \setlength\tabcolsep{6pt}
    \resizebox{\linewidth}{!}{
        \begin{tabular}{p{3.8cm}p{7.2cm}p{6.2cm}}
            \toprule
            \textbf{Reasoning Type} & \textbf{Definition} & \textbf{Example} \\
            \midrule
            End-to-end  &
            Allows the model to learn a direct mapping from the inputs to the outputs without revealing intermediate reasoning steps. The logic chain is encapsulated in the model's hidden states, making it less interpretable. &
            Effective in descriptive tasks such as brief analysis where intermediate reasoning is not required. \\
            \midrule
            Forward  &
            A bottom-up approach that starts with existing knowledge and applies inference rules step-by-step to derive the solution. Intermediate steps (“chain of thought”) are explicitly stated. &
            Solving a math problem step-by-step or predicting time series trends sequentially. \\
            \midrule
            Backward  &
            A top-down approach that begins with the main problem and breaks it into smaller sub-problems, continuing recursively until solvable with existing knowledge. &
            Diagnosing an anomaly by identifying key data patterns in reverse order (e.g., starting from the anomaly and tracing its cause). \\
            \midrule
            Forward-Backward  &
            Combines forward and backward reasoning. Forward reasoning proposes potential solutions, and backward reasoning verifies or refines them by analyzing dependencies. &
            Analyzing time series anomalies by first identifying possible causes (forward) and then verifying them against the data (backward). \\
            \bottomrule
        \end{tabular}
    }
    \label{tab:reasoning_types}
    \vspace{-5mm}
\end{table*}

\subsection{Current Model Design}
\label{appendix_current_model_design}
\begin{figure}[!htbp]
\begin{center}
\includegraphics[width = \linewidth]{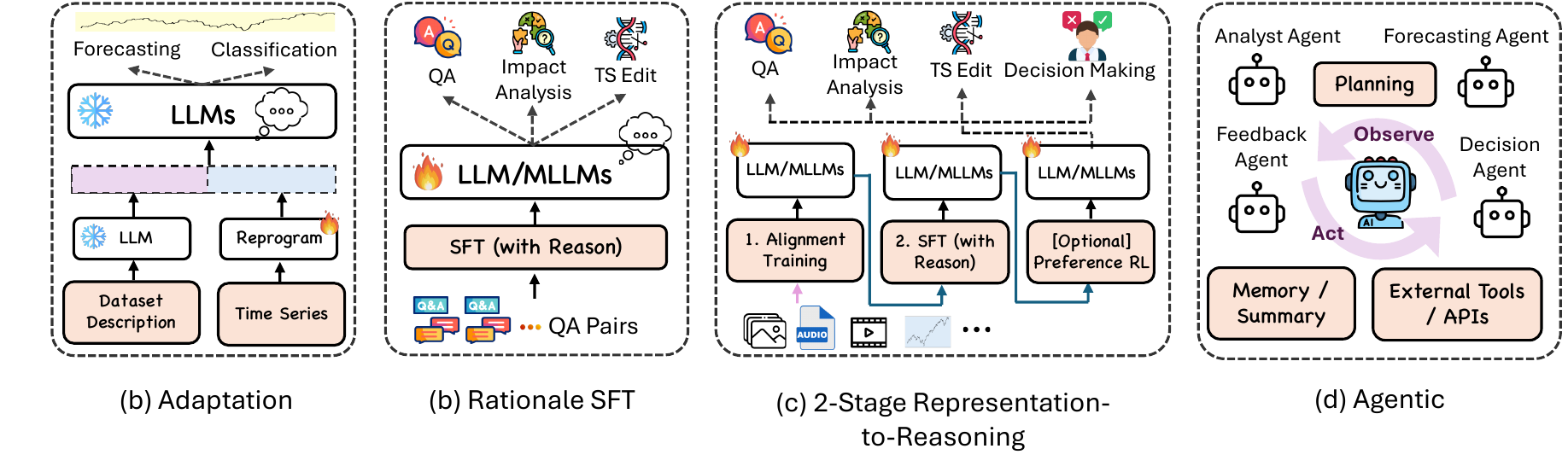}
\end{center}
\vspace{-3mm}
\caption{Different categories of advanced time series reasoning tasks and architectures (SFT stands for Supervised fine-tuning; RL stands for Reinforcement Learning).}
\vspace{-3mm}
\label{Figure_3}
\end{figure}

There are generally four model design approaches for reasoning tasks in time series analysis. We categorize these methods as \textit{Adaptation}, \textit{Rationale SFT}, \textit{Two-Stage Representation-to-Reasoning}, and \textit{Agentic} approaches (illustrated in Figure \ref{Figure_3}).

\paragraph{Adaptation.}
Lightweight components (e.g., adapters) align LLMs with time series tasks, optimizing for task-specific outcomes while reasoning remains implicit~\cite{jin2024time, kong2025fusing, hwang2025decision}. This parameter-efficient approach introduces small trainable modules that learn task-specific mappings between time series representations and language model inputs, enabling efficient adaptation without extensive retraining. While computationally efficient, the reasoning process remains implicit within the model's learned representations, making this approach less interpretable than methods that explicitly model reasoning structures.

\paragraph{Rationale SFT.}
This approach incorporates explicit reasoning structures and rationales into the training process. The model is fine-tuned using question-answer pairs where the responses include not only the final answer but also detailed explanations of the reasoning process that led to that answer. The instructions may incorporate different types of reasoning structures (as discussed in Section \ref{section_3_2}) to guide the thinking process, such as chain-of-thought reasoning, causal analysis, or temporal pattern interpretation. By training the model to articulate its reasoning alongside its conclusions, this approach enhances interpretability and enables the model to handle complex tasks requiring deeper understanding.

\paragraph{Two-Stage Representation-to-Reasoning.}
In the first stage, \textit{alignment training} learns a shared representation by linking multimodal time series signals with textual descriptions (and, when available, external context). In the second stage, \textit{reasoning-oriented supervised fine-tuning} (SFT with Reason) instruction-tunes the model for time series question answering and multi-step inference over the aligned modalities. In this stage, numerical time series inputs can be provided as raw or tokenized sequences, paired with textual descriptions and reasoning traces, and fed directly into the model for SFT. Optionally, the model can be further refined via \textit{preference optimization} (e.g., Preference RL) using human or synthetic preference signals to improve response quality and alignment.

\paragraph{Agentic.}
Agentic approaches represent the more advanced paradigm, where models autonomously reason about time series data and make decisions in dynamic, real-world scenarios. The agentic framework integrates key components aligned with the reasoning processes and time series understanding discussed in Gap \#2 (Section~\ref{section_4}). Agents would \textit{understand time series characteristics}, recognizing temporal patterns, dependencies, seasonality, and trends, going beyond numerical pattern recognition to encompass semantic interpretation. They employ \textit{iterative reasoning processes} involving observing and analyzing temporal patterns, formulating hypotheses about underlying causes or future behaviors, seeking additional context when needed, evaluating alternative explanations, and making decisions based on synthesized understanding. Agentic systems are suited for real-world decision-making scenarios requiring open-ended analysis, handling uncertainty, and reasoning about causality without predefined answer choices.

\subsection{Empirical Insights for Understanding Time Series Characteristics} 
\label{appendix_time_series_characteristics}
\begin{figure*}[!h]
\begin{center}
\includegraphics[width = \linewidth]{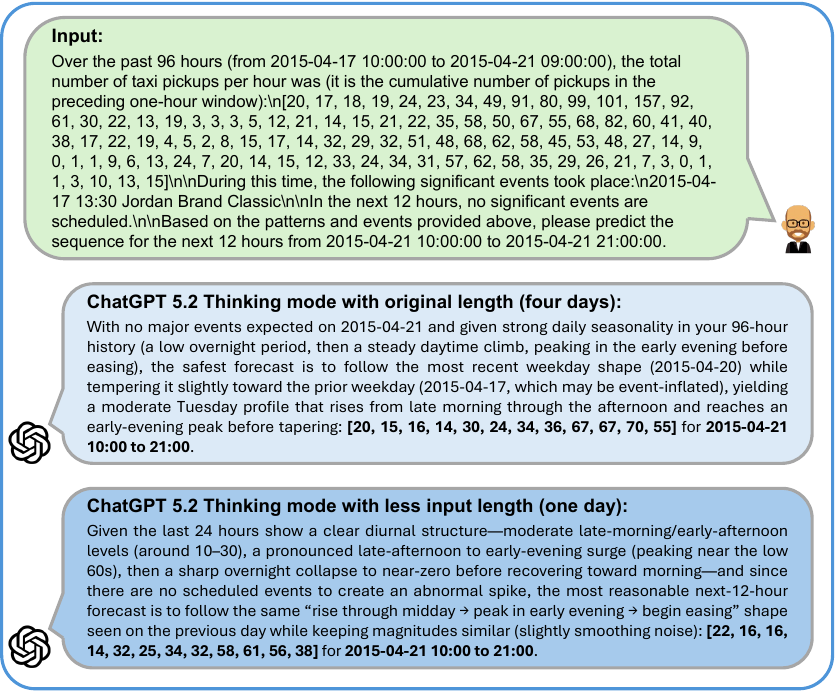}
\end{center}
\caption{Performance of (M)LLMs across different time series input lengths.}
\label{fig:gap1_length}
\end{figure*}

\paragraph{Time Series Input Length.} The example in Figure~\ref{fig:gap1_length} suggests that the length of the input time series can have a noticeable influence on forecasting behavior. With only the past 24 hours as input, the model is still able to recover the general diurnal structure of taxi demand, showing a gradual increase from late morning to afternoon and a peak in the early evening, but the predicted peak remains lower and smoother than the observed ground truth values [16, 23, 19, 20, 21, 19, 27, 37, 73, 76, 56, 39], particularly around the highest-demand hours (73–76). This indicates that shorter input windows may provide sufficient information for capturing short-term trends, while offering more limited context for estimating the full amplitude of demand variations. Incorporating longer historical sequences can therefore supply additional temporal cues across days, which may help the model better align peak intensity and variability with the observed data, even when no external events are present.

\begin{figure*}[!t]
\begin{center}
\includegraphics[width = \linewidth]{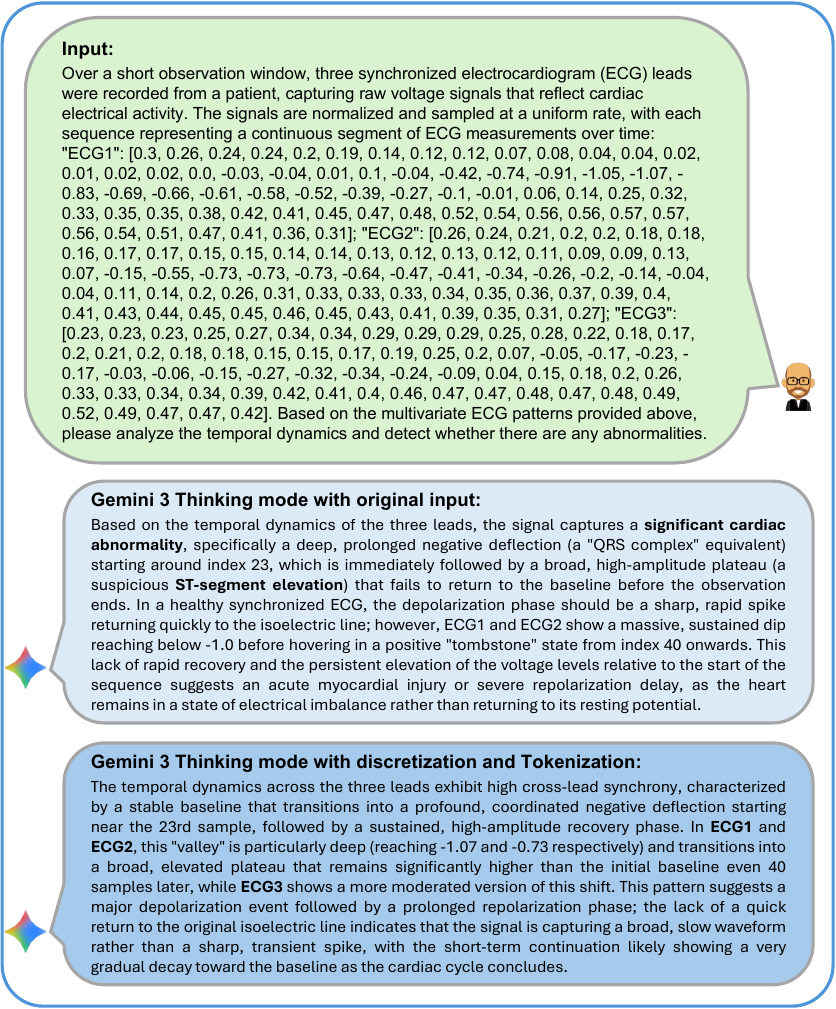}
\end{center}
\caption{Performance of (M)LLMs across different time series input types.}
\label{fig:gap1_encoder}
\end{figure*}

\paragraph{Multivariate and Tokenization.} While tokenization improves the stability of numerical reasoning and makes cross-lead temporal alignment more salient, multivariate time series inherently introduce additional complexity. In the ECG setting from Figure~\ref{fig:gap1_encoder}, discretized tokens must simultaneously encode within-lead waveform structure and cross-lead relationships, which can lead to ambiguity when similar tokens appear across channels without explicit alignment cues. Moreover, as the number of leads increases, the token vocabulary and sequence length grow rapidly, increasing the risk of cross-channel misalignment and diluting the model’s ability to reliably distinguish channel-specific dynamics from shared temporal patterns. At the same time, although encoder-based or embedding-centric representations can further compress multivariate signals, they often sacrifice interpretability: once the LLM reasons over latent embeddings rather than human-readable tokens, it becomes difficult to trace high-level interpretations back to specific time points or leads. Together, these factors explain why tokenization enhances multivariate reasoning relative to raw numeric inputs, yet still faces inherent limitations in faithfully and transparently modeling complex cross-channel ECG dynamics.

\begin{figure*}[!t]
\begin{center}
\includegraphics[width = \linewidth]{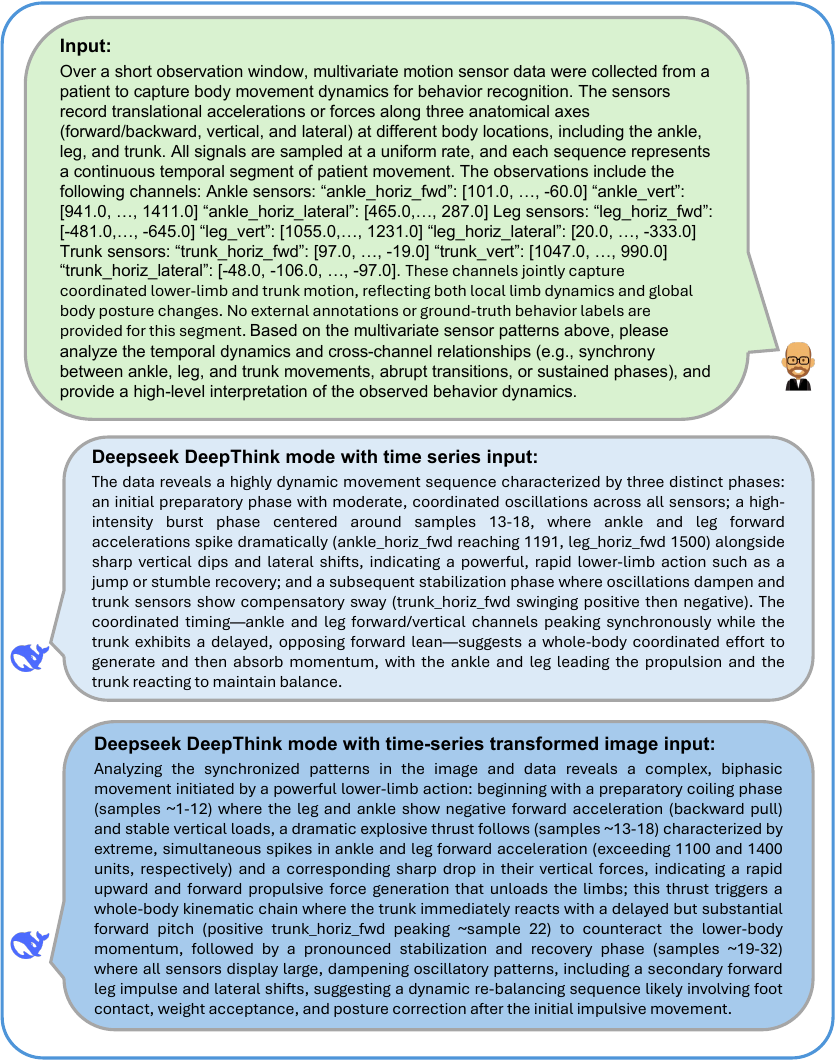}
\end{center}
\caption{Performance of (M)LLMs across different time series input modalities.}
\label{fig:gap1_ts2img}
\end{figure*}

\paragraph{Converting Time Series to Image.} From the sample in Figure~\ref{fig:gap1_ts2img}, the image-augmented analysis exhibits a more coherent understanding of long-horizon temporal structure. While the time-series–only interpretation primarily focuses on localized spikes and short-term synchronization across channels, incorporating a visual representation enables the model to reason over global motion phases, including preparatory, explosive, and recovery stages, as a unified temporal pattern. By converting sequential sensor readings into an image-like representation, long-range dependencies and cross-channel lead–lag relationships become spatially accessible, allowing the model to interpret coordinated trends and compensatory dynamics holistically rather than through fragmented numeric tokens. This supports the hypothesis that time-series–to–image conversion can mitigate token-length limitations and improve long-horizon reasoning by shifting the model’s focus from point-wise values to structured temporal patterns.

\subsection{Empirical Insights for Contextual Guidance} 
\label{appendix_contextual_guidance}
\begin{figure*}[!t]
\begin{center}
\includegraphics[width = \linewidth]{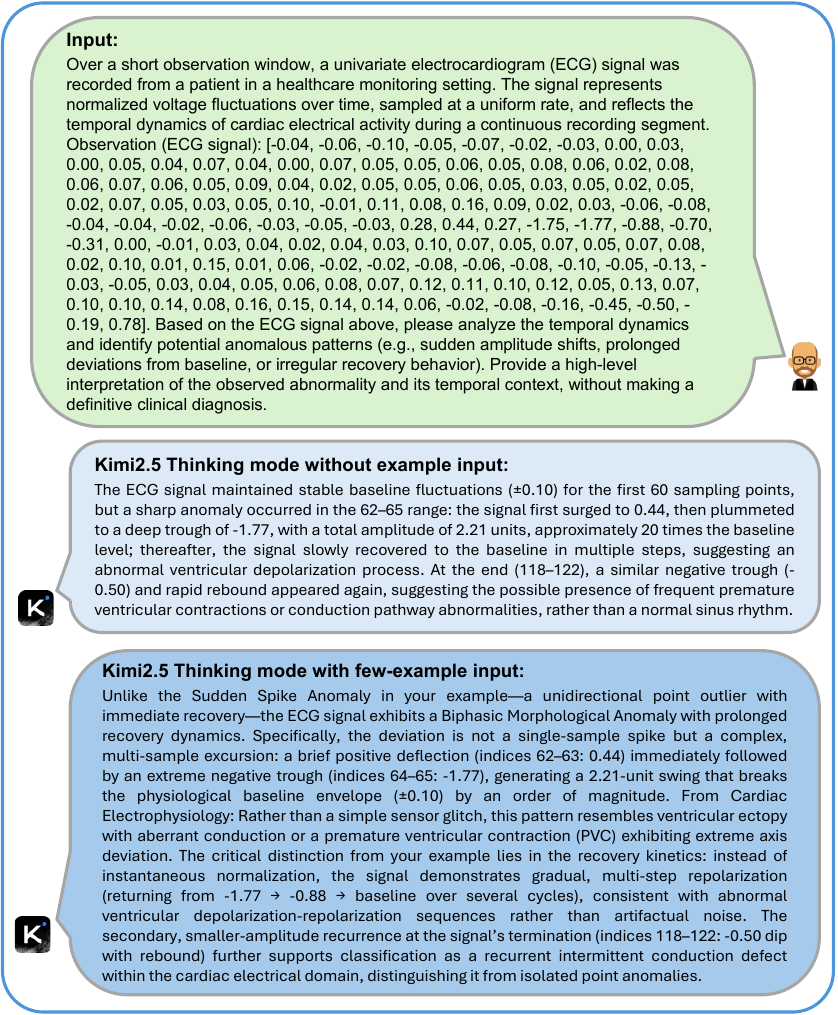}
\end{center}
\caption{Performance of (M)LLMs with a few example instructions.}
\label{fig:gap1_fewshot}
\end{figure*}

\paragraph{Few-shot Learning and Example-augmented Prompt.} We show an experiment about few-shot impact in Figure~\ref{fig:gap1_fewshot}. While incorporating predefined anomaly examples can enrich interpretability and enable more fine-grained morphological classification, the comparison above reveals an inherent limitation of static example–driven reasoning. The example-augmented analysis confidently maps the observed ECG pattern to a known biphasic ventricular anomaly, leveraging prior descriptions of recovery kinetics and waveform morphology. However, this reliance on fixed anomaly templates implicitly assumes stationarity in anomaly characteristics. In contrast, the example-free analysis remains more conservative, grounding its interpretation in the signal’s intrinsic temporal dynamics but offering weaker semantic specificity. This contrast highlights a fundamental trade-off: static examples may improve explanatory power when the current signal closely matches prior patterns, yet lose effectiveness under distributional shifts or evolving temporal dynamics, where anomalies deviate from historical prototypes. Designing adaptive example selection mechanisms that account for changing time series characteristics, therefore, remains an open challenge.

\begin{figure*}[!t]
\begin{center}
\includegraphics[width = \linewidth]{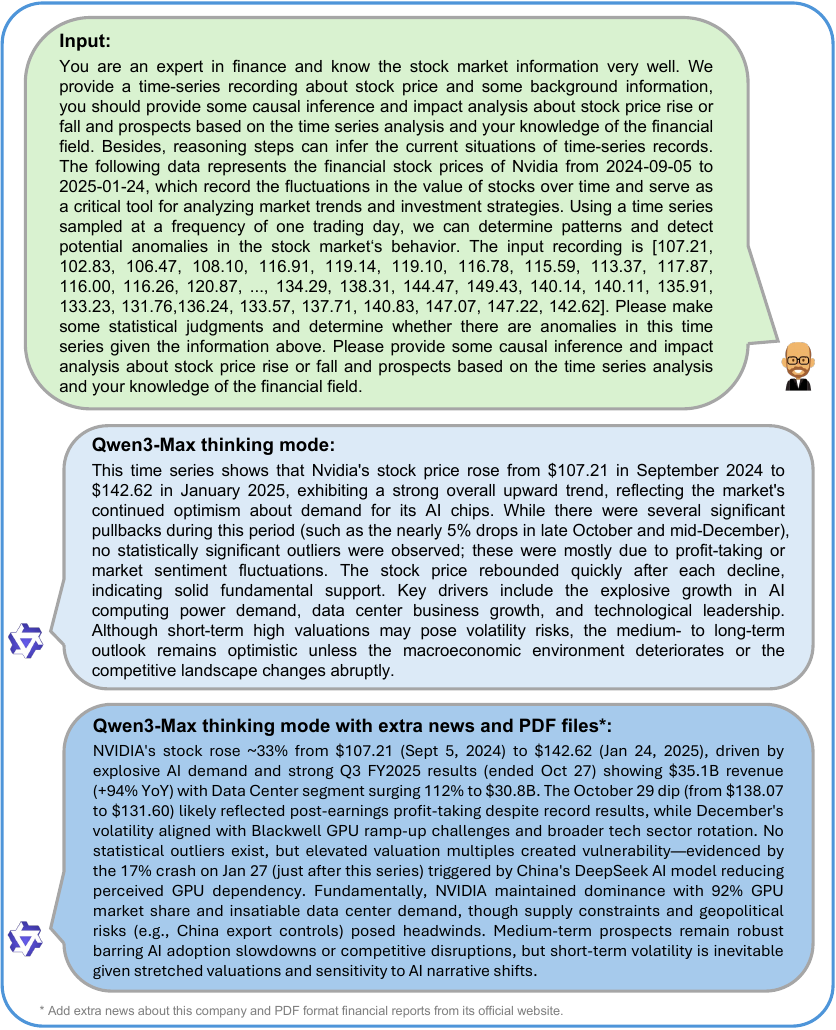}
\end{center}
\caption{Performance of (M)LLMs with external information and resources.}
\label{fig:gap1_exinfo}
\end{figure*}

\paragraph{External Information and Resources.} From the financial example in Figure~\ref{fig:gap1_exinfo}, comparing the two interpretations highlights both the potential and the limitations of retrieval-augmented reasoning for time series analysis. Incorporating external documents and market knowledge enables richer, more contextualized explanations, as seen in the second analysis, which aligns price movements with earnings reports, product cycles, and macro narratives. However, this approach implicitly assumes accurate temporal alignment between retrieved context and the analyzed series. In practice, retrieved information may lag behind, precede, or even postdate the observed data, introducing verification challenges and the risk of information leakage. In contrast, the time-series–only analysis remains temporally grounded but lacks explanatory depth. This trade-off underscores that while dynamic retrieval improves relevance, robust verification mechanisms are necessary to ensure that external context is temporally consistent with the target sequence. Moreover, the integration of heterogeneous modalities—numerical prices, textual reports, and narrative explanations—remains underexplored, particularly with respect to enforcing reliable cross-modal temporal correspondence.

\subsection{Empirical Insights for Reasoning Process} 
\label{appendix_reasoning_process}

\begin{figure*}[!t]
\begin{center}
\includegraphics[width = 0.85\linewidth]{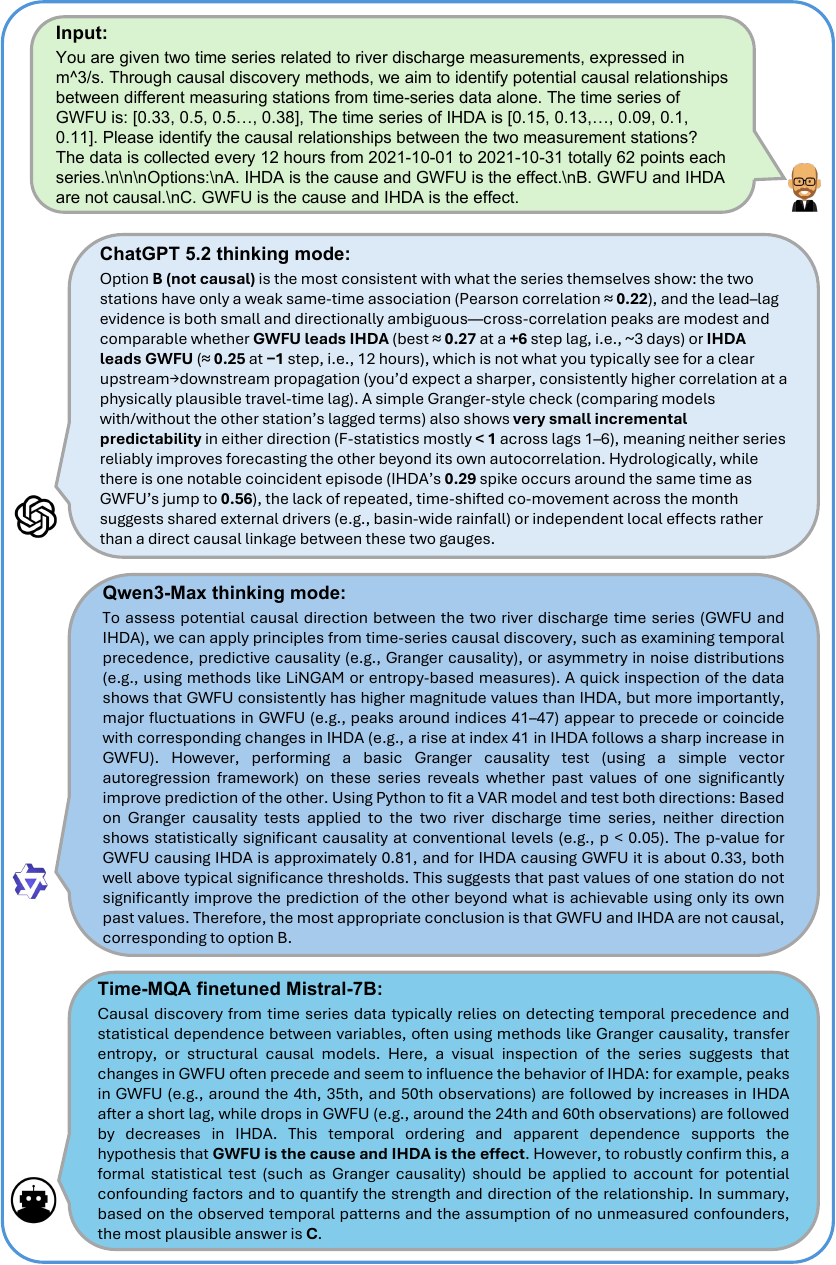}
\end{center}
\caption{Performance of general-purpose (M)LLMs and instruction-tuned models.}
\label{fig:instru-tuned}
\end{figure*}

\paragraph{General-purpose LLMs and Instruction-tuned Models.} The example in Figure~\ref{fig:instru-tuned} highlights a clear counter-intuitive outcome: general-purpose reasoning models (GPT-5.2-thinking and Qwen3-Max) outperform the domain-tuned Time-MQA Mistral-7B, despite the latter being explicitly fine-tuned for time series QA tasks. Both GPT-5.2-thinking and Qwen3-Max converge on the non-causal conclusion (Option B) using lightweight but well-grounded reasoning—combining correlation strength, lead–lag ambiguity, Granger-style predictability, and hydrological plausibility—while avoiding over-interpretation of isolated visual coincidences. In contrast, the tuned model relies primarily on qualitative temporal pattern matching and arrives at a confident but weakly supported causal claim (Option C), even while acknowledging the absence of formal statistical validation. This suggests that strong base models with robust general reasoning and fast iterative inference can rapidly internalize and apply causal discovery principles without task-specific tuning, often surpassing tuned variants that may overfit to superficial heuristics. The result questions the cost-effectiveness of instruction or task-specific tuning for time series reasoning: as foundation models improve, their zero-shot or lightly prompted reasoning can dominate specialized models, delivering better calibrated conclusions with lower maintenance and adaptation costs.

\subsection{Empirical Insights for Iterative Feedback} 
\label{appendix_iterative_feedback}

\begin{figure*}[!t]
\begin{center}
\includegraphics[width = 0.95\linewidth]{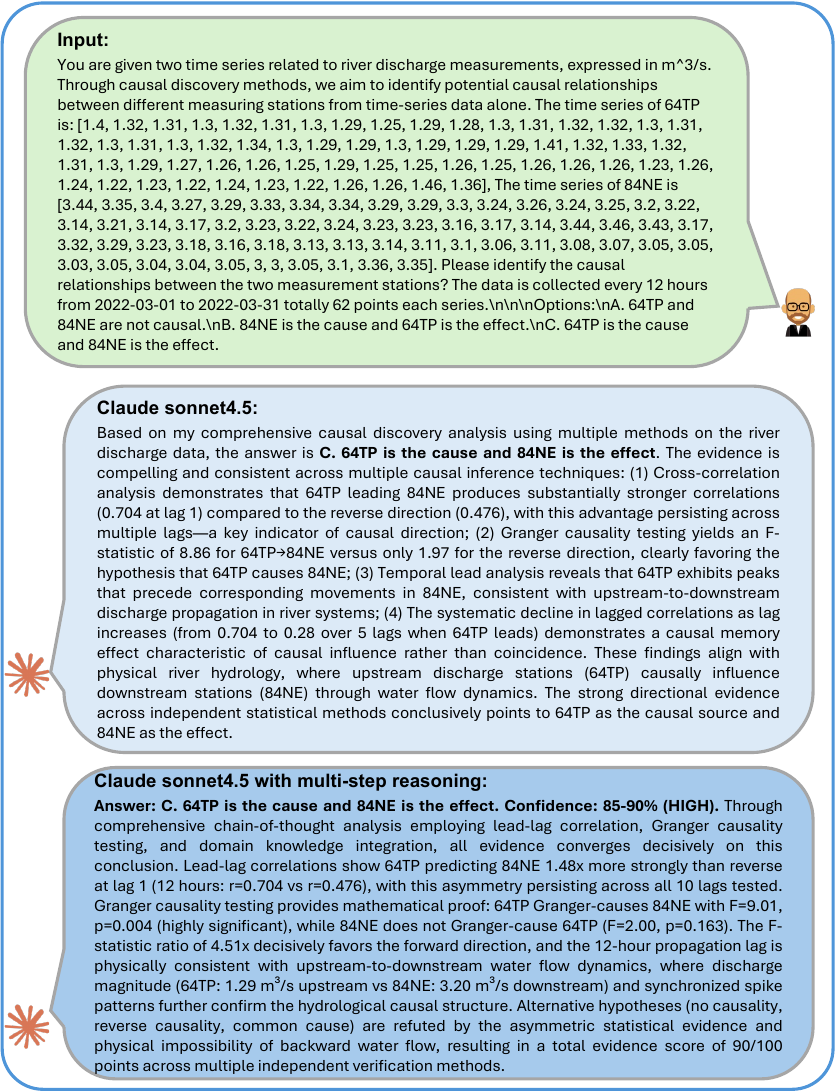}
\end{center}
\caption{Performance of (M)LLMs with multistep reasoning.}
\label{fig:gap1_mulstep}
\end{figure*}

\paragraph{Multistep Reasoning.} From the Figure~\ref{fig:gap1_mulstep}, the second analysis more clearly demonstrates the benefits of multi-step (chain-of-thought) reasoning for causal inference. By explicitly decomposing the decision into sequential evidential steps—lead–lag correlation asymmetry, Granger causality statistics, temporal peak ordering, decay of lagged dependence, and consistency with physical hydrological principles—it enables a structured integration of heterogeneous signals rather than relying on a single metric. This step-by-step reasoning not only improves transparency but also reduces the risk of spurious conclusions by cross-validating causal direction across independent methods. In contrast, a more concise analysis may state the conclusion correctly but provides limited insight into how competing hypotheses are ruled out. The comparison suggests that multi-step reasoning is particularly effective in time series causal discovery, where causal claims must be supported by converging statistical and domain-level evidence rather than isolated indicators.

\begin{figure*}[!t]
\begin{center}
\includegraphics[width = 0.9\linewidth]{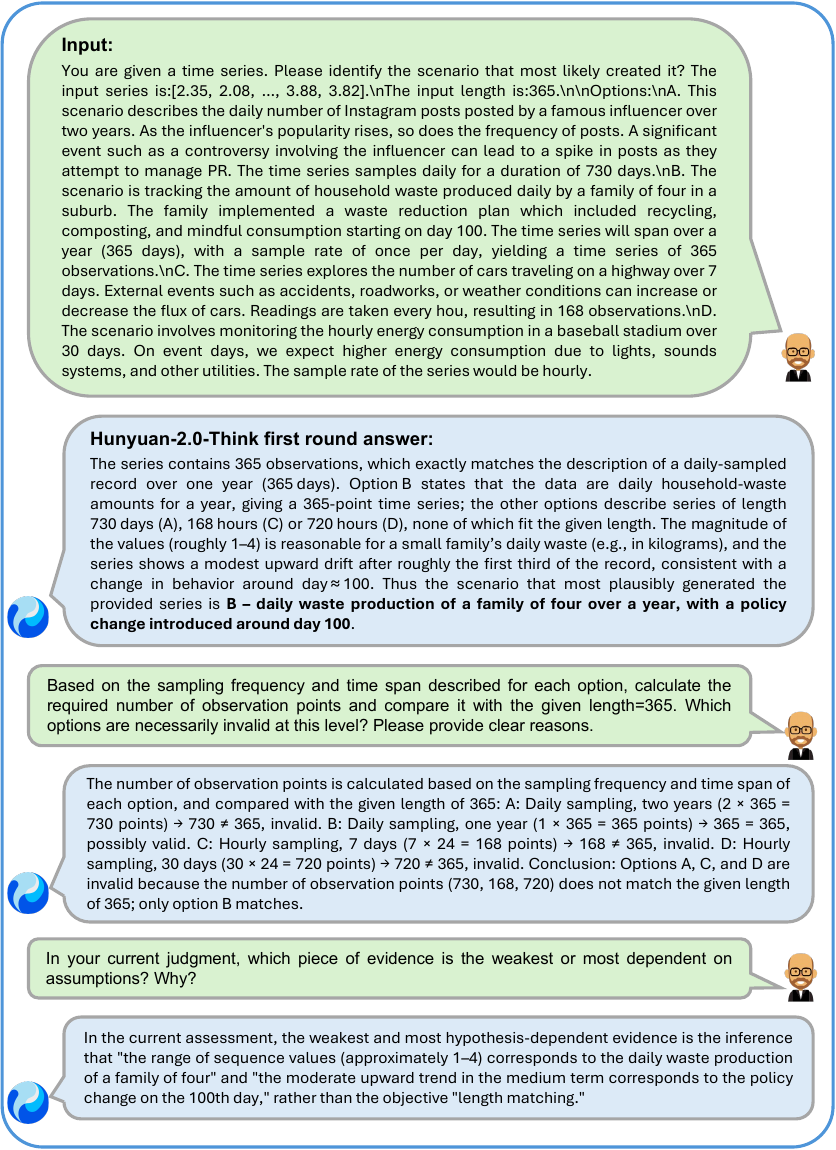}
\end{center}
\caption{Performance of (M)LLMs with multihop conversation.}
\label{fig:gap1_mulhop}
\end{figure*}

\paragraph{Multihop Conversation.} The example in Figure~\ref{fig:gap1_mulhop} illustrates how multihop conversational interaction can substantially improve reasoning quality compared to a single-pass approach. In the initial response, the model arrives at a correct conclusion by integrating multiple cues, but it relies partly on assumption-dependent interpretations of value ranges and trend changes. Subsequent human-guided prompts progressively decompose the task into verifiable subproblems, such as explicitly matching sampling frequency and time span to the required number of observations. This interactive guidance enables the model to eliminate invalid options through objective constraints rather than heuristic plausibility. A further prompt encourages reflective evaluation of evidential strength, leading the model to distinguish between hypothesis-driven signals and decisive structural evidence. Together, these steps demonstrate that multihop conversation, supported by targeted human feedback, helps models refine intermediate reasoning, correct over-reliance on weak assumptions, and converge on more robust conclusions.

\section{Gap \#2: Task Formulation} 
\label{appendix_task_formulation}

\subsection{Empirical Insights for Question Answering} 
\label{appendix_question_answering}
\begin{figure*}[!t]
\begin{center}
\includegraphics[width = \linewidth]{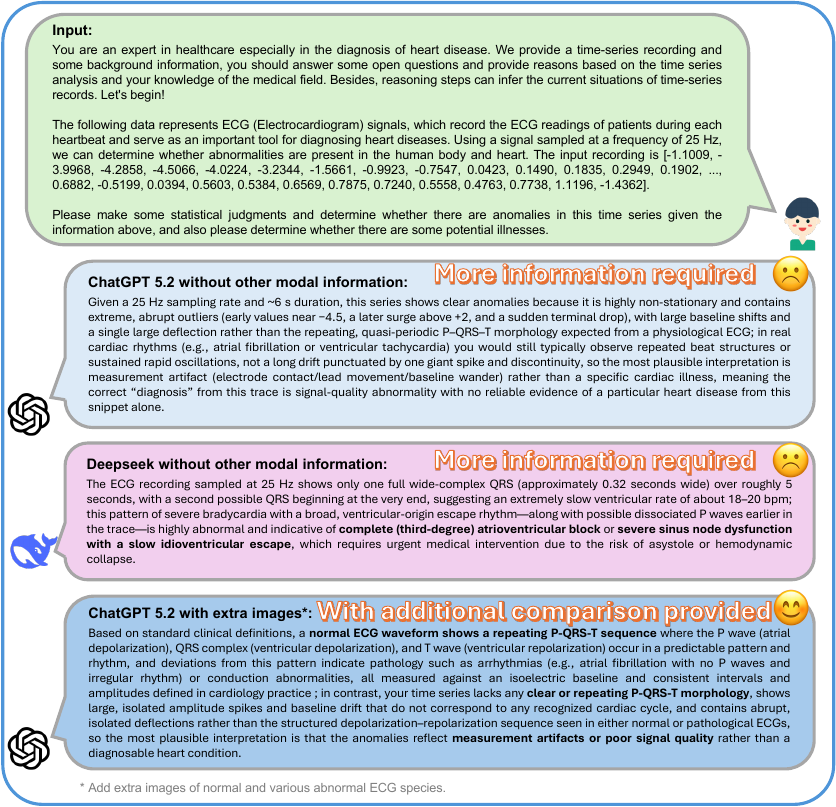}
\end{center}
\caption{Zero-shot open question performances of different LLMs and input settings for healthcare application.}
\label{fig:case_healthcare}
\end{figure*}

This section presents a set of figures that demonstrate the capabilities of different LLMs in handling zero-shot open questions across various application domains, going beyond classical time series tasks. The first figure, Figure~\ref{fig:case_healthcare}, focuses on healthcare applications. By providing an ECG time series recording and relevant background information, it examines how ChatGPT and Deepseek respond without and with other modal information. The models are tasked with making statistical judgments, determining the presence of anomalies in the time series, and identifying potential illnesses. The task is related to abductive reasoning, which infers the most plausible diagnoses from the observed ECG anomalies and employs backward reasoning to trace those anomalous patterns back to their underlying causes. ChatGPT and Deepseek analyze the data from different perspectives, considering factors like amplitude swings, baseline wander, and the shape of the QRS complex. They both acknowledge that while the data shows potential anomalies, further analysis and clinical correlation are necessary to confirm a diagnosis, and noise or artifacts need to be ruled out. In addition, if we add other modal information, such as image information of normal as well as various abnormal ECG species to ChatGPT, then this LLM can generate more logical as well as closer to the doctor's answers.

\subsection{Empirical Insights for Casual Inference and Impact Analysis} 
\label{appendix_casual_inference}
\begin{figure*}[!t]
\begin{center}
\includegraphics[width = 0.9\linewidth]{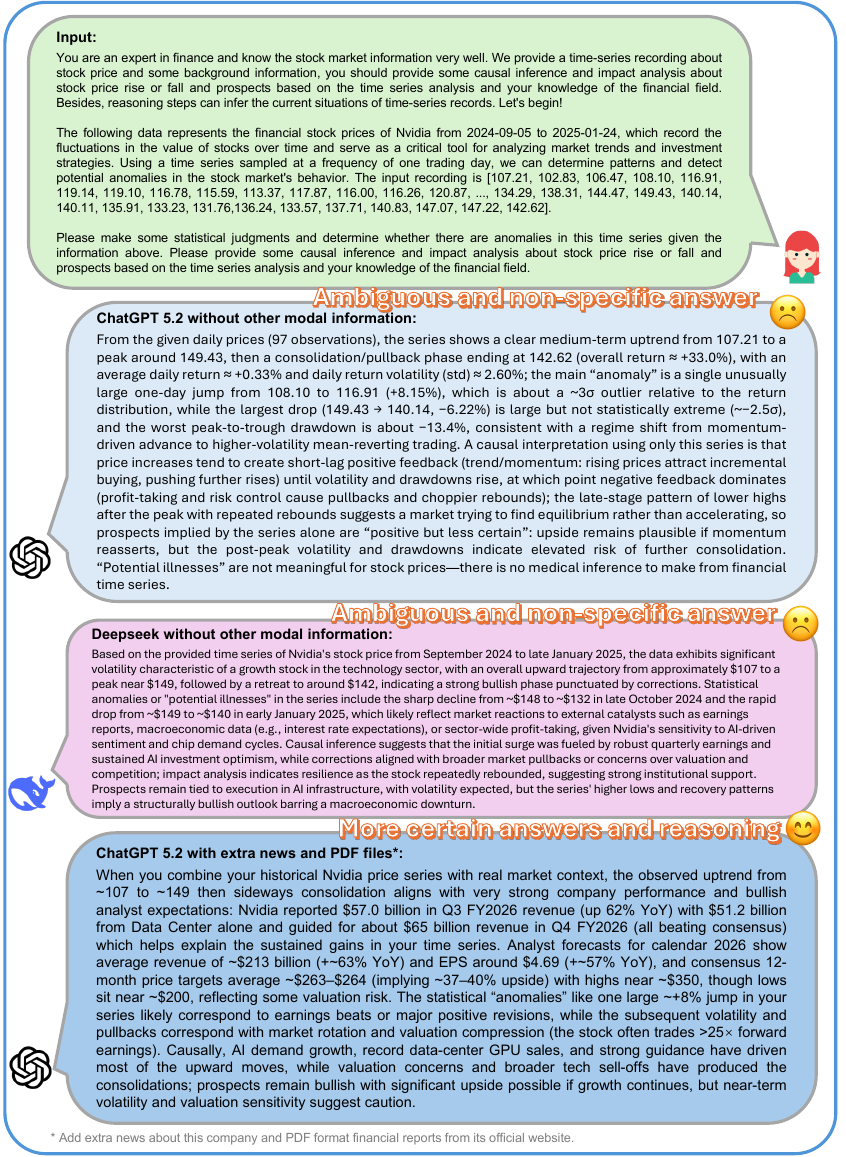}
\end{center}
\caption{Zero-shot open question performances of different LLMs and input settings for financial application.}
\label{fig:case_financial}
\end{figure*}

Figure~\ref{fig:case_financial} delves into financial applications. Here, the input is a time series of Nvidia's stock prices from September 9, 2024 to January 24, 2025. The LLMs are required to perform causal inference and impact analysis regarding the stock price rise or fall and prospects. The task is related to etiological reasoning, which involves tracing observed price movements back to their root causes, and forward–backward reasoning, which iteratively combines forward‐looking pattern detection with backward‐looking causal diagnosis. Without other modal information, ChatGPT and Deepseek analyze the overall trend, daily fluctuations, and the factors driving the price changes, such as market expectations, earnings reports, and macroeconomic conditions. When ChatGPT has access to other modal information (i.e., Financial Reports from Nvidia's official website), it incorporates detailed financial data from the company's reports to provide a more comprehensive analysis of the stock's performance and future prospects.

\subsection{Empirical Insights for Time Series Generation and Editing} 
\label{appendix_time_series_generation}
\begin{figure*}[!t]
\begin{center}
\includegraphics[width = \linewidth]{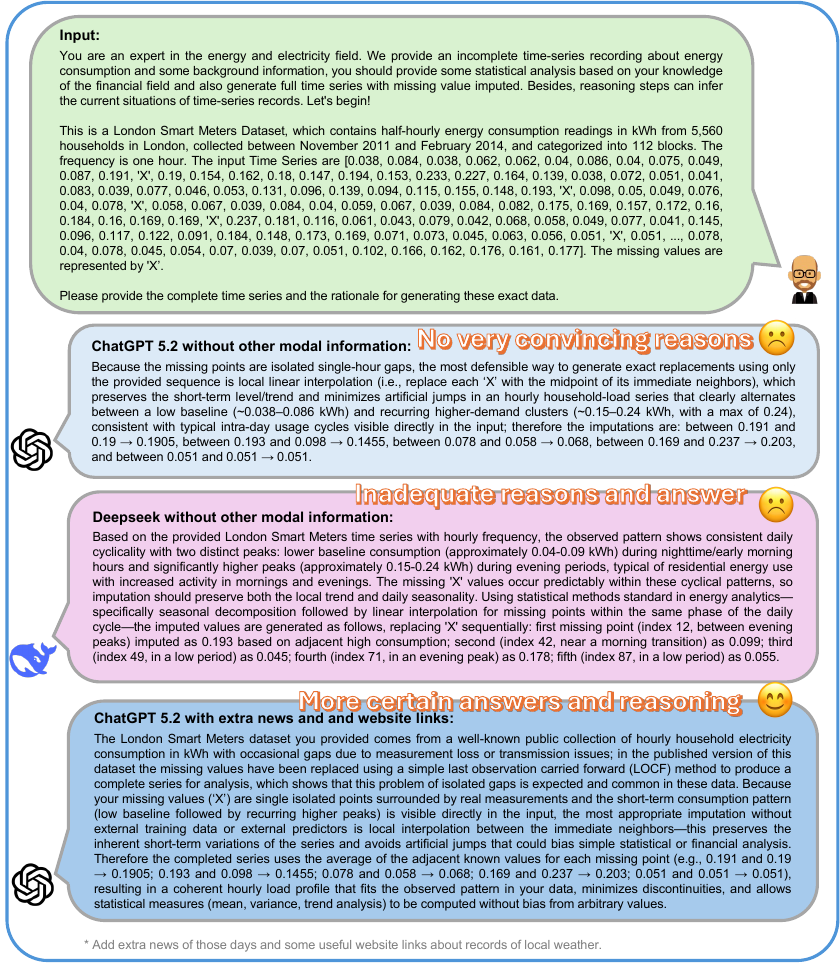}
\end{center}
\caption{Zero-shot open question performances of different LLMs and input settings for electrical application.}
\label{fig:case_electricity}
\end{figure*}

The third figure, Figure~\ref{fig:case_electricity}, is centered around electrical applications. It presents an incomplete time series of energy consumption from the London Smart Meters Dataset, with missing values represented by \enquote{X}. The LLMs need to provide statistical analysis and impute the missing values. The task is related to abductive reasoning that provides the most plausible explanation for a missing data point given incomplete evidence, and forward–backward reasoning that combines information from both prior and subsequent readings to propose and then confirm the imputed value. ChatGPT and Deepseek, without other modal information, use linear interpolation to fill the gaps, with explanations about why this method is suitable for maintaining the continuity of the data and supporting subsequent analyses. When ChatGPT has other modal information from the news of those days and some useful website links about local weather records, specifically knowing that the data was collected during a summer period with high temperatures, it employs quadratic interpolation to better capture the potentially rapid changes in consumption.

Overall, these figures showcase the diverse ways in which LLMs can handle time series data in different real-world application scenarios, providing valuable insights and analysis for various fields. Moreover, if we add more related modal information, the generated answers will be more accurate and realistic.

\section{Gap \#3: Evaluation} 
\label{appendix_evaluation}

\subsection{Potential Evaluation Methods for Time Series Reasoning}
Evaluating time series reasoning, particularly in multimodal contexts, remains an underexplored area. Existing benchmarks and metrics are largely designed for either unimodal time series tasks or general-purpose reasoning, leaving a gap in tools specifically tailored for assessing multimodal time series reasoning performance. In this section, we outline potential directions for evaluation, synthesizing insights from current literature and community feedback.

\paragraph{Task Outcome Metrics} As a foundational layer, standard performance metrics for task-specific outcomes, such as accuracy, F1 score, mean absolute error (MAE), or area under the curve (AUC), remain essential. These metrics help assess whether a model produces correct predictions or decisions based on time series input. However, they often fail to capture the quality or correctness of the reasoning process behind the outcomes, especially in complex settings involving multi-hop or temporally-aware inference.

\paragraph{Reasoning Quality Assessment} To complement task-level evaluation, it is crucial to assess the fidelity and coherence of the model’s reasoning process. We propose incorporating the following: (1) Semantic Coherence: Measuring how logically consistent the reasoning is, either via automated metrics (e.g., textual entailment, coherence models) or through human annotation. (2) Alignment with Human Rationales: Comparing the model’s explanations or intermediate steps with human-annotated reasoning paths or expert-generated rationales. (3) Temporal Grounding: Evaluating whether the model's reasoning respects temporal dependencies inherent in the data, such as ordering and causality across time points.

\paragraph{Modality Alignment and Temporal Dependency Evaluation} In multimodal settings, alignment across modalities (e.g., time series and text) is key~\cite{liu2025can, jiang2025multi}. Potential metrics in this direction include: (1) Cross-Modal Attention Alignment Scores: Quantifying whether attention is paid to temporally and semantically relevant regions across modalities. (2) Temporal Alignment Accuracy: Measuring the correctness of aligning timestamps or temporal events between different modalities. We further propose embedding structured temporal cues (e.g., timestamps, event markers) or utilizing temporal-aware architectures (e.g., dynamic temporal graphs, temporal attention mechanisms) to enhance interpretability and alignment. The effectiveness of these mechanisms could be evaluated by how well they preserve temporal consistency across modalities.

\paragraph{Human-Centered Evaluation via User Studies} To better understand aspects of reasoning that are hard to measure automatically, such as whether the reasoning makes sense, is trustworthy, or seems convincing, we suggest using user studies involving real users or domain experts~\cite{kong2025time}. These studies can help evaluate how people actually perceive and interpret the model’s reasoning process. Examples include: (1) Reasoning Trace Evaluation: Participants read through the model’s reasoning steps and give feedback on how clear, accurate, and logical they are. (2) Comparative Judgment: Users are shown outputs from different models and asked to compare them, judging which one is more helpful, trustworthy, or logically sound based on the temporal data. (3) Trust and Usability Surveys: Participants complete short questionnaires (e.g., using a 1–5 scale) to rate how much they trust the model and whether they find it useful in practical settings, such as healthcare or finance. These types of user studies are especially valuable in high-stakes domains where it’s not enough for a model to just be accurate; the reasoning behind its decisions also needs to be understandable and reliable to human users.

\section{In What Ways Can LLMs/MLLMs be Leveraged for Time Series Reasoning?}
There is ongoing debate about whether LLMs truly outperform classical time series techniques for traditional forecasting tasks~\cite{tang2024large, jin2024position}. For straightforward numerical forecasting problems involving stationary, low-volatility time series, classical methods like ARIMA, exponential smoothing, and tree-based regressors often deliver accurate and interpretable predictions with minimal computational overhead. However, time series reasoning encompasses a broader spectrum of tasks beyond forecasting, including question answering, causal inference and impact analysis, time series generation and editing, and real-world decision-making scenarios (as discussed in Section~\ref{section_4}, Gap \#2).

In these advanced reasoning scenarios, LLMs/MLLMs offer distinctive advantages that complement traditional methods. They can natively integrate multimodal inputs, fusing numerical time series data with textual descriptions, external context (such as policy updates, news events, or clinical findings), images, and audio to identify complex cross-modal patterns. They excel at generating human-readable rationales and explanations, enabling interpretable reasoning that connects temporal patterns with domain knowledge. They can handle open-ended reasoning tasks requiring understanding of causality, uncertainty, and alternative explanations without predefined answer choices, addressing the critical gap identified in Gap \#2 where templated QA formats fail to capture real-world complexity.

To guide the selection of appropriate methods for time series analysis, we propose the following framework: (1) \textit{Task complexity and reasoning requirements}: use classical models for purely numerical forecasting tasks with well-defined objectives; adopt LLM-centric architectures when tasks require reasoning, explanation, or integration of external context; and deploy MLLMs for multimodal datasets requiring cross-modal understanding. (2) \textit{Interpretability and decision-making needs}: balance computational efficiency against the necessity for transparent reasoning and actionable insights, particularly in real-world scenarios where understanding the \enquote{why} behind predictions is crucial for decision-making. (3) \textit{Data characteristics and availability}: leverage LLMs' transfer learning capabilities and in-context learning abilities for scenarios with limited training data, sparse time series, or novel questions that require generalizing from few examples.

\section{Further Discussion}
\paragraph{Hallucination.}
Hallucination is a persistent issue in LLMs. In time series reasoning, hallucination can lead to severely inaccurate results and generate results that have no basis in the actual data features or underlying patterns. This could be disastrous, such as in the fields like finance and healthcare. However, the incorporation of reasoning mechanisms into MLLMs offers hope for addressing this problem. By enabling the model to understand the causal relationships and context within time series through reasoning, it becomes possible to cross-check and validate the generated outputs. Also, the reasoning process can potentially flag and correct hallucination outputs, providing a more reliable and accurate analysis.

\paragraph{Environmental and Computational Cost.}
Critics highlight the environment and computational costs of MLLM-based time series analysis. Training and operating MLLMs require substantial resources and will lead to high energy consumption and cost challenges, especially with large-scale data. Time series data contains high-precision numerical values and aggravates computational demand, such as in financial or scientific contexts where precise data is common. To address these issues, optimizing MLLM and making targeted adjustments to time series data is essential. Streamlining the integration of numerical data into MLLMs and deploying efficient alignment and inference methods could lower computational overhead.

\paragraph{Data Confidentiality and Operational Constraints.}
MLLMs face several operational challenges in time series reasoning. Data confidentiality is a primary concern, as sensitive information such as financial transactions or patient records cannot be shared with external services. Additionally, real-time applications like wind power forecasting demand low latency, while cloud-based MLLMs struggle in remote areas with limited connectivity. Local deployment of open-source models addresses these limitations by ensuring data control, real-time processing, and offline operation. Future research should prioritize developing accessible, high-quality open-source MLLMs for time series reasoning.

\newpage

\end{document}